\pdfoutput=1
\documentclass[11pt]{article}
\usepackage[final]{acl}

\usepackage{times}
\usepackage{latexsym}
\usepackage[T1]{fontenc}
\usepackage[utf8]{inputenc}
\usepackage{microtype}
\usepackage{inconsolata}
\usepackage{graphicx}

\usepackage{amsmath, amssymb}
\usepackage{hyperref}
\usepackage{subcaption} 
\usepackage[dvipsnames,table, svgnames]{xcolor}
\usepackage[table]{xcolor}  
\usepackage{xcolor}  
\newcommand{\bad}[1]{\cellcolor{MediumSlateBlue!20}{#1}}
\newcommand{\good}[1]{\cellcolor{red!20}{#1}} 
\usepackage{multirow} 
\usepackage{float}
\usepackage{makecell}
\usepackage{multirow}

\usepackage[most]{tcolorbox}
\newtcolorbox{myquote}[1][]{%
    colback=black!3,
    colframe=black!3,
    notitle,
    sharp corners,
    borderline west={2pt}{0pt}{blue!80!black},
    enhanced,
    breakable,
}
\usepackage{fontawesome5}
\usepackage{xcolor}
\newcommand{\cmark}{\textcolor{green}{\faCheckCircle}} 
\newcommand{\xmark}{\textcolor{red}{\faTimesCircle}}   
\usepackage{enumitem}

\usepackage[table]{xcolor}
\definecolor{dualnoexp}{HTML}{64ddd2}   
\definecolor{dualexp}{HTML}{183f9f}     
\definecolor{KV}{HTML}{15a6b1}     
\definecolor{IV}{HTML}{ff6234}     

\usepackage{epsfig}
\usepackage{graphicx}
\usepackage{booktabs}
\usepackage{makecell}
\usepackage{lipsum}
\usepackage{tabularx}
\usepackage{diagbox}
\usepackage[export]{adjustbox}
\usepackage{pifont}
\usepackage{url}
\usepackage{tcolorbox}
\usepackage{color}
\usepackage{amsmath}
\usepackage{amssymb}
\usepackage{multirow, varwidth}
\usepackage{dblfloatfix}
\usepackage{verbatim}
\makeatletter
\newif\if@restonecol
\makeatother

\usepackage[ruled,vlined]{algorithm2e}
\usepackage{xr}
\usepackage[amsmath,thmmarks]{ntheorem}
\usepackage{xspace}

\DeclareRobustCommand\onedot{\futurelet\@let@token\@onedot}
\def\onedot{. }

\author{
Chuyi Kong$^{1}$,~ 
Wei Gao$^{2}$,~
Jing Ma$^{1}$\thanks{The corresponding author. Code is available on \href{https://github.com/renatz/REFLEX}{GitHub}.},~
Hongzhan Lin$^{1}$,~
Yuxi Sun$^{1}$\\
$^{1}$ Hong Kong Baptist University \quad
$^{2}$ Singapore Management University \\
{\texttt{kongchuyi@life.hkbu.edu.hk} \quad \texttt{weigao@smu.edu.sg} \quad \texttt{majing@hkbu.edu.hk}}\\
}

\title{%
\raisebox{-0.4\height}{%
  \includegraphics[width=1cm]{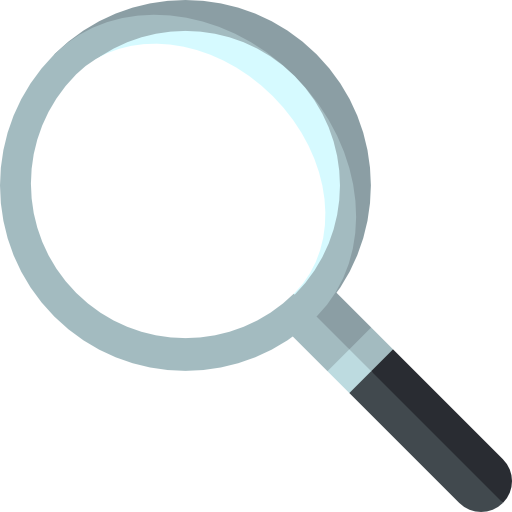}%
}\hspace{0.6em}%
\begin{tabular}[t]{@{}l@{}}
REFLEX: Self-Refining Explainable Fact-Checking via\\
~~~~~~~~~~~~~~~~~~~Verdict-Anchored Style Control
\end{tabular}
}

\begin{document}

\maketitle
\begin{abstract}
The prevalence of fake news on social media calls for automated fact-checking systems that deliver not only accurate verdicts but also faithful explanations. However, existing large language model (LLM)-based methods often overlook deceptive misinformation styles in generated explanations, producing unfaithful rationales that may mislead human judgment. They also rely heavily on external knowledge sources, which can introduce hallucinations and incur substantial latency, undermining both reliability and responsiveness in real-time settings. To address these limitations, we propose \textbf{RE}ason-guided \textbf{F}act-checking with \textbf{L}atent \textbf{EX}planations (\textsc{REFLEX}), a self-refining framework that explicitly controls reasoning style by anchoring explanations to the predicted verdict. \textsc{REFLEX} leverages self-disagreement veracity signals between a backbone model and its fine-tuned variant to construct steering vectors, thereby naturally disentangling factual content from stylistic cues. Experiments on a real-world benchmark show that \textsc{REFLEX} achieves state-of-the-art performance under LLaMA-series models using only 465 self-refined samples. Owing to its transferability, \textsc{REFLEX} also yields gains of up to 7.54 Macro-F1 points on in-the-wild data. Further analysis shows that our method effectively mitigates faithful hallucination, leading to both more reliable explanations and more accurate verdicts than prior explainable fact-checking approaches.

\end{abstract}

\section{Introduction}

\begin{figure}[htbp]
\centering
\includegraphics[width=\linewidth]{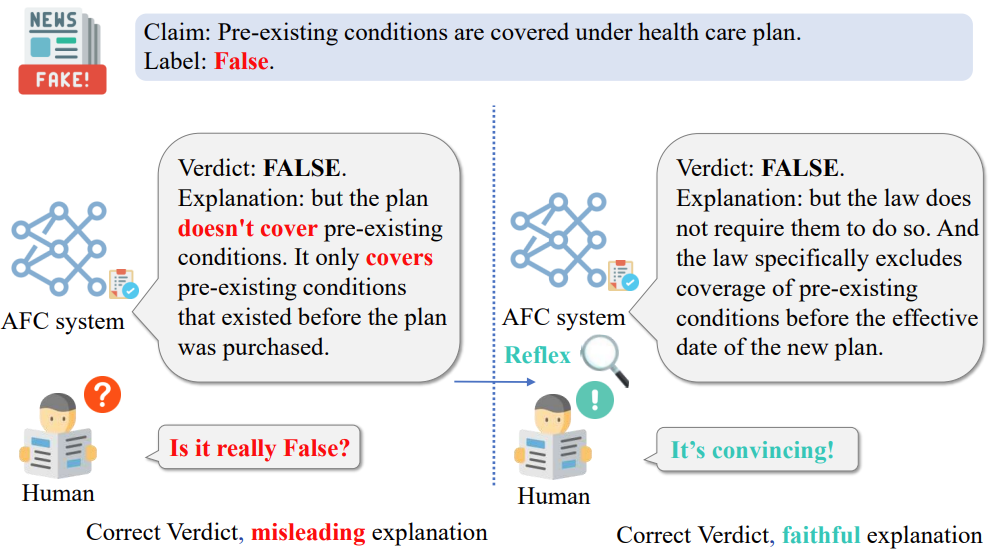} 
\caption{A motivating case showing how deceptive explanatory style can mislead human judgment.}
\vspace{-13pt}
\label{fig:case}
\end{figure}

The rapid spread of fake news on social media poses a major challenge for automatic fact-checking~(AFC)~\cite{guo-etal-2022-survey}. Recent AFC systems increasingly leverage Large Language Models (LLMs) to predict verdicts and generate explanations. Yet they often overlook a core requirement for human-facing fact-checking: explanations must be not only plausible, but also faithful. As shown in Figure~\ref{fig:case}, even when an LLM predicts the correct verdict, an unfaithful explanation can still mislead human judgment.

Such failures often stem from deceptive explanatory styles in LLM-generated rationales~\citep{turpin2023language, chencan}, a phenomenon related to faithfulness hallucination~\cite{hallu-llm}. These rationales may begin with factual statements but gradually drift into verdict inconsistency, internal contradiction, or overconfident fabrication. We therefore argue that effective explainable fact-checking requires disentangling \textbf{\textit{fact}} from deceptive \textbf{\textit{style}} in explanations. In the worst case, such unfaithful explanations may even cascade into incorrect verdicts, since LLMs appear to construct holistic \textit{internal} conceptual representations before producing rationalized outputs, rather than simply mirroring next-token prediction~\cite{anthropic2025tracing}.

\begin{figure*}[htbp!]
\centering
\includegraphics[width=\linewidth]{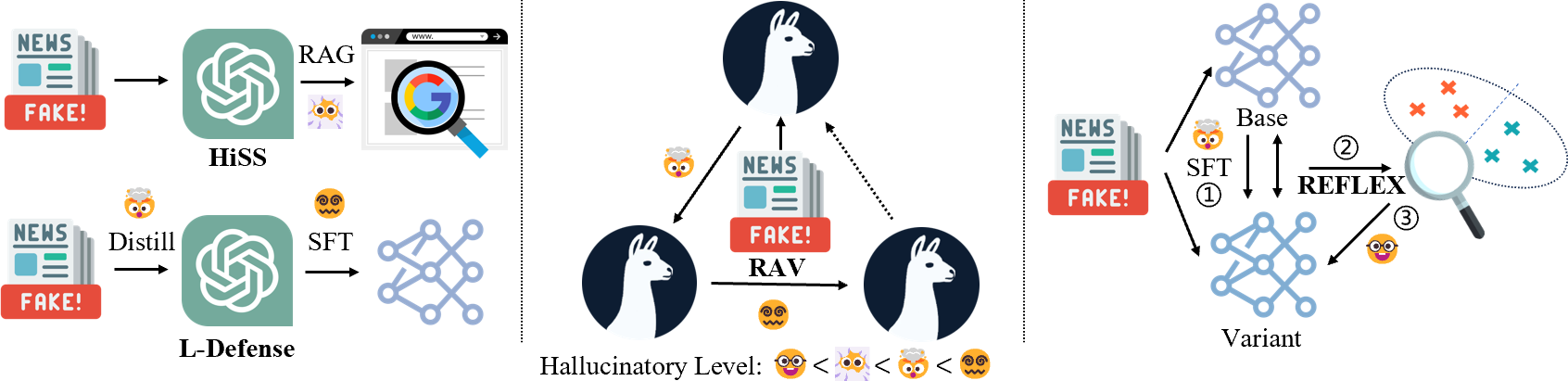} 
\caption{Comparison between REFLEX and representative baselines. Hallucination levels are indicated by emojis. The dashed line denotes an optional process. Bidirectional arrows in REFLEX denote \textbf{self-disagreement} signals.}
\vspace{-16pt}
\label{fig:compar}
\end{figure*}

Existing AFC methods such as HiSS~\cite{zhang2023towards} and L-Defenses~\cite{wang2024explainable} mainly address these issues through \textit{external} interventions. While such methods reduce costly manual involvement, their dependence on external closed-source APIs or search engines introduces additional latency and may inject misinformation or hallucinations~\citep{gekhman2024does} into fine-grained explanations. This compromises both reliability and responsiveness, which are critical in real-world fact-checking.

To address these limitations, we improve explanation faithfulness by steering \textit{internal} model activations. Our approach exploits self-disagreement veracity signals across the model training lifecycle, from a backbone model to its post-trained variants, to disentangle transferable Steering Vectors~(SVs)~\cite{han2023word} into Inference Vectors~(IVs) and Knowledge Vectors~(KVs). IVs capture style-sensitive veracity gains that emerge after fine-tuning. Under limited supervision, fine-tuning is often more effective at shaping task-specific reasoning styles and activating latent backbone knowledge~\cite{li2023inference} than at injecting new factual knowledge~\citep{ghosal2024understanding,ren2024learn,zhao2025style}. Amplifying these signals therefore helps align explanations with verdicts. In contrast, KVs capture fact-sensitive veracity drops caused by knowledge conflicts~\citep{gekhman2024does} introduced during fine-tuning, which can induce alignment tax~\cite{Leike_2022} and hallucinations. Suppressing these signals helps preserve consistent factual representations. Based on this insight, we adaptively activate the appropriate vector according to probability gaps and use it to refine explanations, yielding our self-refining AFC paradigm, namely, \textbf{RE}ason-guided \textbf{F}act-checking with \textbf{L}atent \textbf{EX}planations (\textsc{REFLEX}).

Experiments show that our internal-activation-based approach outperforms methods that rely on external resources, either in verdict accuracy alone or jointly in verdict accuracy and explanation quality with more faithful and concise reasoning. Moreover, \textsc{REFLEX} generalizes across backbones, datasets, and limited-sample settings, demonstrating strong transferability (up to 7.54 Macro-F1 points), flexibility (up to 7.57 Macro-F1 points), and data efficiency. We further find that evidence-free training is better suited to our paradigm, as external evidence can introduce noise and amplify hallucinations. In addition, we observe a divergence amplification effect: larger directional discrepancies between KVs and IVs yield more effective disentanglement and stronger gains. Post-hoc analysis further shows that explanations produced by \textsc{REFLEX} are more logically coherent and less misleading, thereby mitigating faithfulness hallucination and improving verdict accuracy.

Overall, our main contributions are as follows:
\begin{itemize}
    \item We propose \textsc{REFLEX}, a simple yet effective self-refining paradigm that disentangles fact from style, improves explanation faithfulness, and mitigates misinformation.
    \item \textsc{REFLEX} achieves state-of-the-art performance on a real-world dataset using only a small number of self-refined training samples, while producing high-quality explanations.
    \item We show that \textsc{REFLEX} supports cross-model transfer and exhibits performance-adaptive transferability, which can reduce response latency in fact-checking.
    \item We identify a divergence amplification effect in \textsc{REFLEX}, where greater directional divergence yields larger performance gains.
\end{itemize}

\section{Background}
\paragraph{Faithfulness in NLP.}
In NLP, faithfulness is typically studied from two perspectives. The first concerns consistency between model predictions and the reasoning that supports them~\citep{jacovi2020towards, atanasova2023faithfulness, parcalabescu2024measuring}. However, recent work~\cite{yan2025phd} suggests that probabilistic models may not capture genuine reasoning, but instead rely on superficial patterns, i.e., \textbf{\textit{style}}. We therefore use human-written explanations as learning targets to better approximate human reasoning style. The second concerns consistency with the surrounding context, often referred to as \textbf{\textit{faithfulness hallucination}}~\cite{hallu-llm}, including instruction, context, and logical inconsistency. Our work follows this second line. Recent fact-checking work~\cite{kim2024can} addresses explanation--evidence alignment with multi-agent systems; in contrast, we adopt an evidence-free single-model setting to avoid hallucinations introduced by external evidence and iterative interaction.

\paragraph{Explainable Fact-Checking}

\begin{figure*}[htbp!]
\centering
\includegraphics[width=\linewidth]{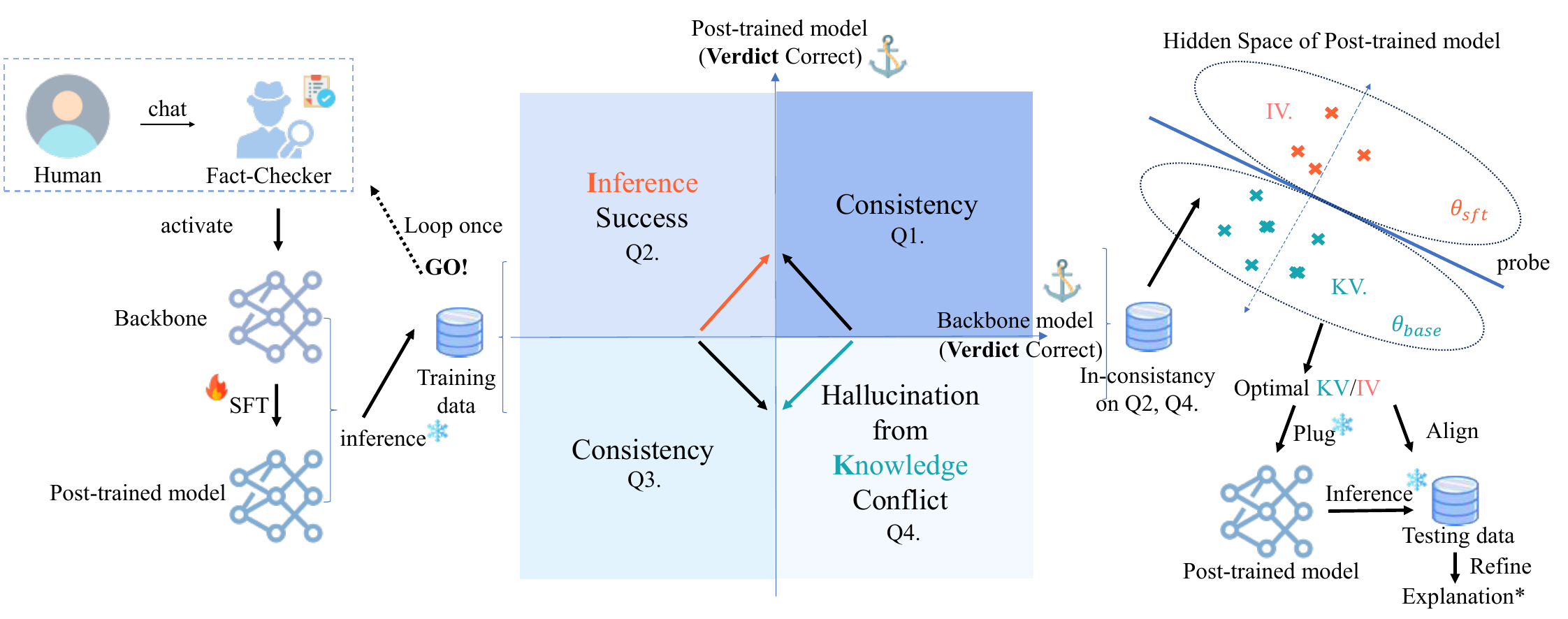} 
\caption{Overview of the three-stage \textsc{REFLEX} paradigm. Red text denotes \textcolor{IV}{reasoning style} acquired during post-training, and green text denotes \textcolor{KV}{distorted factual knowledge} in the backbone model.}
\vspace{-10pt}
\label{fig:mt}
\end{figure*}

With the rise of LLMs, recent work has explored automatic explanation mechanisms to reduce human annotation effort in explainable fact-checking. As shown in Figure~\ref{fig:compar}, HiSS~\cite{zhang2023towards} decomposes complex claims into atomic claims and verifies them with RAG~\cite{rag}, using retrieved reasoning trajectories as explanations. 
L-Defense~\cite{wang2024explainable} distills adversarial evidence and explanations from teacher models. RAV~\cite{shukla2025recon} builds a fully open-source multi-agent dialogue system for claim verification and treats LLM responses as explanations.
Despite their promise, these methods face several \textbf{key challenges}. RAG-based methods may introduce hallucinations at both retrieval and generation stages~\cite{qian2023merge, hu-etal-2025-removal}. Distillation and fine-tuning can amplify the snowballing effect~\cite{Leike_2022} of teacher-model hallucinations. Multi-agent systems are vulnerable to hallucination propagation~\citep{jain2025collaborative,liang2026multi} and incur additional interaction latency, making them less suitable for time-sensitive fact-checking. To address these limitations, \textsc{REFLEX} introduces a self-refining and transferable paradigm that mitigates hallucinations in explanations, reduces inference latency, and improves the interpretability of internal model behavior.

\paragraph{Style Control in Fact-Checking}
\label{sec: scifc}

Several studies improve fact-checking by exploiting stylistic differences across claim sources (e.g., human vs.\ machine) to improve verdict prediction~\citep{perez2018automatic,rashkin2017truth}. However, humans can deliberately manipulate style to deceive, which weakens the reliability of such signals~\citep{schuster2020limitations,wu2024fake}. In contrast, we anchor style control on verdict prediction, steering the model toward less deceptive explanations that remain aligned with the verdict.

Our approach builds on a broader line of work on style control, ranging from controllable text generation~\citep{dathathriplug, li2022diffusion} to activation editing~\citep{liemergent,hernandez2023inspecting}. Recent steering-vector methods also follow a contrastive paradigm. ITI~\cite{li2023inference} identifies truth-related directions at inference time, while CCS~\cite{burnsdiscovering} and CAA~\cite{sv-llama2} learn contrastive directions associated with truthfulness or stylistic behavior. However, in explainable fact-checking, \textbf{style and fact are entangled}: longer explanations and their associated verdicts may diverge due to attention dispersion~\cite{wei2025shadows}, making strict contrastive pairing infeasible. To address this issue, we refine explanations through cross-stage sample selection and by decomposing steering vectors into KVs and IVs. To avoid shortcut learning over factual signals alone, we further adopt label-free interventions.

\section{Methodology}

\subsection{Task Formulation}

Given a dataset $\mathcal{D}=\{(c, evi, v, exp)_i\}_{i=1}^N$, where $c$ denotes a claim, $evi$ optional evidence, $v$ the gold veracity label, and $exp$ a human-written explanation, \textsc{REFLEX} aims to generate verdicts and explanations such that the explanation remains faithful to the predicted verdict, thereby mitigating faithfulness hallucination.

As shown in Figure~\ref{fig:mt}, \textsc{REFLEX} consists of three stages:
(1) \textbf{Dialogue-style Fact-Checker Training}, which reformulates fact-checking as a dialogue task and trains LLMs to jointly generate verdicts and explanations;
(2) \textbf{Adaptive Sample Selection}, which identifies self-disagreement samples between a backbone model and its fine-tuned variants;
(3) \textbf{Self-Explanation Guided Steering (S-EGS)}, which disentangles steering vectors into KVs and IVs from these samples and explicitly aligns explanations with verdicts.

\subsection{Dialogue-style Fact-Checker Training}
\label{sec: mt}

\paragraph{Data Preprocessing.}
\label{sec: dp}

LLM backbones already encode substantial factual knowledge~\cite{li2023inference}, and fine-tuning with limited data often mainly activates this knowledge while adapting the model to the downstream task~\citep{berglund2023taken,ghosal2024understanding,ren2024learn}. We therefore reformulate conventional document-style supervision as a single-turn QA-style dialogue to better activate backbone knowledge. Prior work shows that QA-style supervision yields stronger knowledge generalization during fine-tuning, whereas document-style supervision (e.g., Wikipedia text) yields weaker generalization~\citep{zhao2025style}.

\paragraph{Training Protocol.}
Following prior dialogue-based training, we optimize a standard cross-entropy objective to jointly predict verdicts and explanations. For a comprehensive study, we consider four input-output configurations:

{\small
\[
\begin{aligned}
x=[c]&\to y=[v], & x=[c]&\to y=[v;exp],\\
x=[c;\,evi]&\to y=[v], & x=[c;\,evi]&\to y=[v;exp].
\end{aligned}
\]
}
To encourage coherent reasoning, we incorporate Chain-of-Thought (CoT)~\citep{wei2022chain}, prompting the model to output its reasoning path as the explanation, which has been shown to improve performance~\citep{lippmann2025style}. We further adopt role-play prompting to strengthen reasoning ability~\citep{kong2023better}. Prompt templates and training details are provided in Appendix~\ref{app: pt} and \ref{app: td}.

\subsection{Adaptive Sample Selection}

\paragraph{Self-Knowledge Extraction.}

After training, we run inference on the training set using both the backbone and post-trained models. Since the backbone lacks instruction-following ability~\citep{liu2022generated}, we use few-shot prompting to elicit its knowledge. To avoid data leakage, few-shot examples are drawn either from another training split or from a held-out validation set. To mitigate majority-label bias in few-shot learning, we balance labels when filling the maximum context window. For reproducibility, we fix the temperature to zero.

\paragraph{Cross-Stage Sample Selection.}

Given veracity predictions $\hat{v}^{\text{base}}$ and $\hat{v}^{\text{sft}}$ on the training set, we categorize samples relative to the gold label $v^{\text{gold}}$ as follows:
\[
\begin{alignedat}{2}
\textbf{Quadrant 2:}~~
&\hat{v}^{\text{base}} \neq v^{\text{gold}}, ~\hat{v}^{\text{sft}} = v^{\text{gold}}
\\[-1pt]
&\Rightarrow \textbf{Reasoning Gain},\\[4pt]
\textbf{Quadrant 4:}~~
&\hat{v}^{\text{base}} = v^{\text{gold}},~\hat{v}^{\text{sft}} \neq v^{\text{gold}}
\\[-1pt]
&\Rightarrow \textbf{Knowledge Loss}.
\end{alignedat}
\]
As shown in Figure~\ref{fig:mt}, samples in Quadrant 2 (Q2) capture cases in which post-training corrects the backbone’s verdict. We treat these cases as \textbf{reasoning gains}, because under limited supervision, fine-tuning is more likely to activate latent backbone knowledge and shape task-specific reasoning style than to inject substantial new factual knowledge~\cite{li2023inference,berglund2023taken,ghosal2024understanding,ren2024learn}. In other words, when the post-trained model succeeds where the backbone fails, we attribute the improvement primarily to better reasoning behavior rather than to newly acquired facts.
By contrast, samples in Quadrant 4 (Q4) capture cases in which post-training flips a previously correct backbone verdict into an incorrect one. We interpret these cases as \textbf{knowledge loss}, since the degradation is more plausibly explained by conflicting knowledge or distorted factual representations introduced during post-training, which can lead to alignment tax~\cite{Leike_2022} and hallucinations~\citep{gekhman2024does,huang2025survey}. We therefore adaptively select samples from Q2 and Q4 for subsequent steering.

\subsection{Self-Explanation Guided Steering}
\label{sec:egs}

\paragraph{Disentangled Steering.}
Unlike ITI~\cite{li2023inference} and CAA~\cite{sv-llama2}, which control a single factual or stylistic direction using explicit contrastive pairs and yield one steering vector, we decompose steering vectors into KVs and IVs across the quadrants above to disentangle fact and style. Specifically, we amplify style-sensitive signals by steering the model toward the positive IV direction, thereby aligning explanations with verdicts. Conversely, we suppress fact-sensitive signals by steering the model away from negative KV directions, thereby preserving consistent factual representations. We select the optimal KV or IV by maximizing the probability gap between unsteered and steered outputs, which we find to correlate with verdict accuracy. For each candidate direction, we measure how much steering increases the model’s probability assigned to the gold verdict relative to the unsteered baseline, and select the direction with the largest improvement (details in Appendix~\ref{sec: alignment}).
In addition, because entangled or inconsistent factual and stylistic signals weaken contrastive supervision, we adopt label-free interventions that do not rely on explicitly specified factual directions, enabling more robust in-the-wild steering. To improve conceptual separability while keeping computational overhead low, we use logistic probes and intervene only at decoder blocks. Implementation details are provided in Appendix~\ref{app: td}.

\paragraph{Explanation Refinement.}

To further align explanations with verdicts, we quantify the alignment between each token’s hidden representation and the optimal direction using cosine similarity. For layer $l$ and token $t$, let $h$ denote the hidden state and $\mathbf{s}$ the optimal vector. We define the alignment score as $a_{l,t}=\frac{h_{l,t}\cdot\mathbf{s}_l}{\|h_{l,t}\|\|\mathbf{s}_l\|}$.
Positive scores indicate alignment with the optimal direction, while negative ones indicate the opposite. Segments with dense negative scores are often associated with redundant or noisy stylistic content. We therefore suppress them using the lightweight Ratcliff-Obershelp pattern-matching algorithm~\cite{ratcliff1988} to improve explanation readability.

\section{Experiments and Results}

\begin{table*}[t] 
\centering
\setlength{\tabcolsep}{3pt} 
\renewcommand{\arraystretch}{1.2} 
\resizebox{\textwidth}{!}{%
\begin{tabular}{@{}llll|ccc|ccc@{}}
\toprule
\multirow{2}{*}{Model} & \multirow{2}{*}{x $\to$ y} & \multirow{2}{*}{Model} 
& \multirow{2}{*}{\shortstack{No Closed-Source API Dependency\\ / \# Distilled Explanations for Training}} 
& \multicolumn{3}{c|}{RAW-FC} & \multicolumn{3}{c}{LIAR-RAW} \\
\cmidrule(lr){5-7}\cmidrule(lr){8-10}
& & & & P & R & macF1 & P & R & macF1 \\ \midrule

\multicolumn{3}{c}{\textbf{External Dependency Design}} \\ 
\multirow{2}{*}{ChatGPT}          & c$\to$v; exp & --  & \xmark / -- & 47.72 & 48.62 & 44.43 & 25.41 & 27.33 & 25.11 \\
                 & c, evi$\to$v; exp & --  & \xmark / -- & 39.48 & 45.07 & 39.31 & 29.64 & 23.57 & 21.90 \\
HiSS$_{\text{Google}}$  & c$\to$v; exp& ChatGPT & \xmark / -- &53.40 & 54.50 &53.90&46.80&31.30&37.50 \\
FactLLaMA$_{\text{Google}}$ & c; evi$\to$v & LLaMA2-7B & \xmark / -- & 56.11 & 55.50 & 55.65 & 32.46 & 32.05 & 30.44 \\
L-Defense &c; evi$\to$v; exp & ChatGPT + Roberta-Large  & \xmark / 32,240 & 61.72 & 61.01 & 61.20 & 30.55 & 32.20 & 30.53 \\ 
RAV              & c; evi$\to$v; exp & LLaMA-3.1-70B-Instruct $\times$3 & \cmark / -- & -- & -- & 59.19 & -- & -- & 25.40 \\ \midrule
\multicolumn{3}{c}{\textbf{Internal Control Design~(Ours)}} \\ 
SFT          & c$\to$v; exp & LLaMA2-7B & \cmark / 0 & 60.66 & 61.04 & 60.59 & \textcolor{gray}{48.38} & \textcolor{gray}{46.83} & \textcolor{gray}{43.05} \\ 
S-EGS            & c$\to$v; exp & LLaMA2-7B & \cmark / 465 & \textbf{65.04} & \textbf{65.01} & \textbf{64.99} & \textbf{\textcolor{gray}{52.45}} & \textbf{\textcolor{gray}{50.39}} & \textbf{\textcolor{gray}{50.59}} \\ 

\bottomrule
\end{tabular}%
}
\caption{Performance comparison on RAW-FC and LIAR-RAW. ``Google'' indicates the use of the Google Search API. Full baseline details are provided in Appendix~\ref{app: compute-time}.}
\label{tab:ba}
\end{table*}

\begin{table}[ht]
\centering
\resizebox{\linewidth}{!}{%
\begin{tabular}{l|cccc|cccc}
\hline
                  & \multicolumn{4}{c|}{\textbf{RAW-FC}} & \multicolumn{4}{c}{\textbf{LIAR-RAW}} \\ \cline{2-9} 
                  & \textbf{M$\downarrow $} & \textbf{I} & \textbf{S} & \textbf{R} & \textbf{M$\downarrow $} & \textbf{I} & \textbf{S} & \textbf{R} \\ \hline
Oracle - skyline            & 1.52 & 4.46 & 4.73 & 4.72 & 1.85 & 4.44 & 4.60 & 4.69 \\ 
ChatGPT$_\text{w/ evi}$        & 2.07 & 4.44 & 4.62 & 4.69 & 2.29 & 3.71 & 4.04 & 3.99 \\
ChatGPT$_\text{w/o evi}$       & 1.97 & 4.00 & 4.44 & 4.68 & 2.27 & 3.93 & 4.29 & 4.50 \\ 
L-Defense$_\text{LLaMA2}$   & 1.95 & 4.44 & 4.67 & 4.62 & 2.20 & 4.39 & 4.64 & 4.63 \\
L-Defense$_\text{ChatGPT}$  & 1.91 & 4.17 & 4.41 & 4.49 & 2.06 & 4.12 & 4.28 & 4.47 \\ \hline
\multicolumn{9}{l}{\textbf{Ours}} \\ 
SFT          & 1.90 & 4.78 & 4.82 & 4.55 & 1.90 & 4.48 & 4.60 & 4.65  \\ 
S-EGS$_\text{LLaMA2}$     & \textbf{1.79} & \textbf{4.88} & \textbf{4.83} & \textbf{4.80} & \textbf{1.77} & \textbf{4.58} & \textbf{4.66} & \textbf{4.83} \\ \hline
\end{tabular}%
}
\caption{Automatic evaluation of explanation quality. The best \textsc{REFLEX} variant with LLaMA-2 is reported.}
\vspace{-15pt}
\label{tab:auto_eval}
\end{table}

%
\paragraph{Datasets.}

To better reflect real-world fact-checking and reduce hallucination risk, we use three datasets whose claims originate from professional fact-checking platforms and whose explanations are human-written: RAW-FC~\citep{yang2022coarse}, LIAR-RAW~\citep{yang2022coarse}, and AVeriTeC~\citep{schlichtkrull2023averitec}. In RAW-FC and LIAR-RAW, explanations directly justify the claim label, whereas in AVeriTeC they justify both the claim and its supporting evidence. For simplicity, we refer to all such rationales as explanations (details in Appendix~\ref{app: data}).

\paragraph{Metrics.}
For verdict prediction, we report Precision, Recall, and Macro-F1. For explanation quality, we conduct both automatic and manual evaluations. 
\textbf{(1) Automatic evaluation.} Following \citet{wang2024explainable}, we use ChatGPT as a judge~\cite{gu2025surveyllmasajudge} along four dimensions: \textbf{M}isleadingness, \textbf{I}nformativeness, \textbf{S}oundness, and \textbf{R}eadability. Misleadingness measures whether an explanation is consistent with the veracity label, while Soundness measures whether the explanation is logically valid and grounded; these two dimensions most directly reflect context and logical consistency under faithfulness hallucination. The remaining dimensions are described in Appendix~\ref{app: eval_details}. 
\textbf{(2) Manual evaluation.} For baselines, we follow~\citet{wang2024explainable} and conduct point-wise evaluation on a subset of RAW-FC. For ablations, we perform three rounds of pairwise evaluation on misleadingness over all samples in Section~\ref{sec: sc}.



\subsection{Baseline Trials}

\subsubsection{Baselines}

We compare against two categories of methods:
\textbf{(1) External Dependency Design}: ChatGPT~\cite{chatgpt}, HiSS~\cite{zhang2023towards}, FactLLaMA~\cite{cheung2023factllama}, L-Defense~\cite{wang2024explainable}, and RAV~\cite{shukla2025recon};
\textbf{(2) Internal Control Design}: \textsc{REFLEX} (ours).
Among these, FactLLaMA also uses the Google API to retrieve evidence and fine-tunes LLaMA-2-7B. To remain comparable to most baselines, we adopt LLaMA-2-7B as the backbone in this section.

\subsubsection{Results}

\begin{table*}[]
\centering
\resizebox{\textwidth}{!}{
\begin{tabular}{@{}lllcccccc@{}}
\toprule
Backbone & Stage & x$\to$y & Raw-FC & $\Delta$ mac-F1 & LIAR-RAW & $\Delta$ mac-F1 & AVeriTeC & $\Delta$ mac-F1 \\ \midrule
\multirow{10}{*}{LLaMA-2} & \multirow{4}{*}{BASE} & c$\to$v & \textbf{35.61} & -- & 29.26 & -- & -- & -- \\
 &  & c; evi$\to$v & 27.08 & -- & 16.97 & -- & \textbf{28.18} & -- \\
 &  & c$\to$v; exp$_{\text{cross}}$ & 34.41 & -- & 12.48 & -- & -- & -- \\
 &  & c[; evi]$\to$v; exp$_{\text{self}}$ & 31.68 & -- & \textbf{35.80} & -- & 27.70 & -- \\ \cmidrule(lr){2-9}
 & \multirow{3}{*}{SFT} & c$\to$v & 26.44 & -9.17 & 37.23 & +7.97 & -- & -- \\
 &  & c; evi$\to$v & 44.85 & +17.77 & 40.21 & +23.24 & 75.91 & +47.73 \\
 &  & c[; evi]$\to$v; exp & \textbf{60.59} & +26.18 & \textbf{43.05} & +7.25 & 84.62 & +56.92 \\ \cmidrule(lr){2-9}
 & \multirow{3}{*}{S-EGS} & c$\to$v & 31.47 & \textbf{+5.03} & 38.65 & +1.42 & -- & -- \\
 &  & c$\to$v; exp$_{\text{cross}}$ &  \textbf{64.99} & +4.40 & \textit{42.77} & \underline{-0.28} & -- & -- \\
 &  & c[; evi]$\to$v; exp$_{\text{self}}$ & 61.81 & +1.22 &  43.06 & +0.01 & \textit{84.61} & -0.01 \\ \midrule
\multirow{10}{*}{Qwen-3} & \multirow{4}{*}{BASE} & c$\to$v & 46.54 & -- & 37.63 & -- & -- & -- \\
 &  & c; evi$\to$v & 46.23 & -- & 41.30 & -- & 66.14 & -- \\
 &  & c$\to$v; exp$_{\text{cross}}$ & 46.66 & -- & \textbf{42.25} & -- & -- & -- \\ 
 &  & c[; evi]$\to$v; exp$_{\text{self}}$ & \textbf{48.86} & -- & 39.16 & -- & 66.02 & -- \\
 \cmidrule(lr){2-9}
 & \multirow{3}{*}{SFT} & c$\to$v & 41.67 & -4.87 & 41.72 & +4.09 & -- & -- \\
 &  & c; evi$\to$v & \textbf{63.17} & +16.94 & 42.29 & +1.69 & 85.52 & +19.38 \\
 &  & c[; evi]$\to$v; exp & 58.35 & +9.49 & \textbf{46.73} & +4.48 & 88.02 & +22.22 \\ \cmidrule(lr){2-9}
 & \multirow{3}{*}{S-EGS} & c$\to$v & 41.69 & +0.02 & 41.73 & +0.01 & -- & -- \\
 &  & c$\to$v; exp$_{\text{cross}}$ & \textbf{59.39} & +1.04 & \textbf{47.13} & +0.40 & -- & -- \\
 &  & c[; evi]$\to$v; exp$_{\text{self}}$ & 58.86 & +0.51 & \textit{46.53} & \underline{-0.20} & \textbf{88.21} & +0.19 \\ \bottomrule
\end{tabular}
}
\caption{Macro-F1 of S-EGS across backbones, training stages, and datasets. \textit{cross} denotes few-shot examples drawn from another dataset, while \textit{self} denotes examples drawn from the model’s own validation set. [;] indicates optional evidence inputs, used only for AVeriTeC since its explanations are evidence-grounded. For SFT, $\Delta$ Macro-F1 is computed against the best BASE configuration.}
\vspace{-5pt}
\label{tab: bb}
\end{table*}

For \textbf{verdict prediction}, Table~\ref{tab:ba} shows that our method achieves state-of-the-art performance on RAW-FC without relying on closed-source APIs, retrieved evidence, or large-scale multi-agent systems. After Stage~1 (SFT), it surpasses ChatGPT by 16.16 Macro-F1 points and HiSS by 6.69 points. Compared with FactLLaMA, which uses the same backbone but lacks dialogue-style supervision and full-parameter tuning, our model gains 4.94 points. It performs comparably to RAV while using only a single model, and remains slightly below L-Defense. After applying S-EGS, our method surpasses RAV by 5.80 points and L-Defense by 3.79 points, despite using only 465 self-extracted samples versus 32,240 GPT-3.5-distilled explanations in L-Defense. This highlights the strong data efficiency of \textsc{REFLEX}. Since our models are trained under the three-way label setting used by explanation-based methods, we do not report a direct verdict comparison on LIAR-RAW here.

For \textbf{explanation quality}, we evaluate open-source baselines that generate explanations. Following \citet{wang2024explainable}, we map the six LIAR-RAW labels to three for all applicable baselines, and include an Oracle setting (ChatGPT with gold claims and verdicts) as a skyline. As shown in Table~\ref{tab:auto_eval}, after SFT our model achieves the best performance on RAW-FC in misleadingness, informativeness, and soundness, trailing only L-Defense (ChatGPT-distilled) in readability. On LIAR-RAW, it achieves the best informativeness and readability, and ranks second to L-Defense (LLaMA-2) in soundness. Applying S-EGS further improves all metrics consistently. Human evaluation (Appendix~\ref{app: he}) confirms the same trend. Because LLM-as-a-Judge may exhibit length bias~\citep{gu2025surveyllmasajudge}, we also report explanation lengths in Appendix~\ref{app: el}. Under the same backbone, our explanations are shorter than those of L-Defense on RAW-FC, and shorter than all baselines, including Oracle, on LIAR-RAW, suggesting that \textsc{REFLEX} produces explanations that are both concise and faithful.

\subsection{Ablation Studies}
We conduct the following ablation experiments on top of SFT models as baselines.

\subsubsection{On Backbone Models}
\label{sec:obb}
To assess generalization across backbones, in addition to LLaMA-2-7B, we also train and evaluate Qwen-3-8B~\cite{yang2025qwen3}.

As shown in Table~\ref{tab: bb}, S-EGS improves over SFT on all three datasets in most settings, with gains of up to 5.03 Macro-F1 points, demonstrating strong generalization across backbones. The two marginal drops ($>$ 0.1) are likely due to severe \textbf{recency bias} in few-shot prompting (Appendix~\ref{app: ld}), despite existing mitigation strategies~\citep{lu2022fantastically,min2022noisy,nguyen2023context}. Two patterns are noteworthy: \textbf{(1)} LLaMA-2 benefits more from S-EGS than Qwen-3; \textbf{(2)} AVeriTeC shows the smallest gains, despite the strongest SFT performance and the lowest hallucination rate (Appendix~\ref{app: hr}). We attribute this \textbf{saturation phenomenon} to its QA-style formulation, which already aligns well with our training paradigm (Section~\ref{sec: dp}).

Two additional trends emerge from the training configurations.
\textbf{(1) Using explanations as targets is consistently beneficial.} At the SFT stage, explanation-based supervision improves performance across settings. In contrast, at the BASE stage, LLaMA-2 performs better without explanations, likely because its pretraining already strongly emphasizes factual knowledge~\cite{touvron2023llama}. 
\textbf{(2) Evidence conditioning is often harmful.} At the BASE stage, adding evidence generally degrades performance, especially for LLaMA-2, which is consistent with its weaker intrinsic reasoning ability~\cite{gandhi2025cognitive}. At the SFT stage, excluding AVeriTeC because its explanations are inherently evidence-grounded, we find that although evidence can help under the $c \rightarrow v$ setting, models trained with explanations but without evidence ($c \rightarrow v; exp$) generally outperform models trained with evidence but without explanations ($c; evi \rightarrow v$), except for Qwen-3 on RAW-FC. We conjecture that evidence introduces noise that can amplify hallucinations, while Qwen-3’s stronger reasoning ability partially compensates for this effect~\cite{gandhi2025cognitive}. 
Based on these findings, we use the \textbf{evidence-free} explanation-based objective ($c \rightarrow v; exp$) for RAW-FC and LIAR-RAW in subsequent experiments.

\begin{table}[t]
\centering
\small
\setlength{\tabcolsep}{4.5pt}
\begin{tabular}{l c c c c c}
\toprule
\multirow[c]{2}{*}{Model} & \multirow[c]{2}{*}{Transfer}  & \multicolumn{2}{c}{Source} & \multicolumn{2}{c}{Target} \\
\cmidrule(lr){3-4} \cmidrule(lr){5-6}
      &          & F & $\Delta$ & F & $\Delta$ \\
\midrule
\multirow{2}{*}{LLaMA-2}
 & R $\rightarrow$ L & \underline{64.99} & +4.40 & 50.59 & \textbf{+7.54} \\
 & L $\rightarrow$ R & \underline{43.06} & +0.01 & 47.20 & -13.39 \\
\midrule
\multirow{2}{*}{Qwen-3}
 & R $\rightarrow$ L & \underline{59.39} & +1.04 & 45.12 & -1.61 \\
 & L $\rightarrow$ R & \underline{47.13} & +0.40 & 41.04 & \textbf{-17.31} \\
\bottomrule
\end{tabular}
\caption{Cross-dataset transfer results (Macro-F1). R: RAW-FC. L: LIAR-RAW. $\Delta$: gain over SFT. Corr($\Delta$)=0.89; Corr(Source F, Target $\Delta$)=0.95 (Pearson). See Appendix~\ref{app: full-cmtr} for full results.}
\label{tab: cross-model}
\vspace{-15pt}
\end{table}

\subsubsection{On Cross-model Transfer}

Although LLMs often overfit across domains, and some prior work, e.g., LIMA~\cite{zhou2023lima} and Character-LLM~\cite{shao2023character}, even exploits deliberate overfitting for broader adaptation, our goal is to avoid overfitting and preserve scientific validity under realistic transfer. We therefore conduct cross-model steering using optimal vectors learned on LIAR-RAW and RAW-FC, excluding AVeriTeC because its dialogue format differs substantially from the other two. We also remove overlapping few-shot samples to avoid leakage.

As shown in Table~\ref{tab: cross-model}, we observe two key patterns. 
(1) Transfer performance depends strongly on source model strength: the Pearson correlation between source Macro-F1 and target $\Delta$ Macro-F1 is 0.95. Strong vectors extracted from a better-performing source model (source Macro-F1 = 64.99) can substantially improve a weaker target setting (source Macro-F1 = 43.06) by up to 7.54 points, whereas weak vectors extracted from a weaker model (source Macro-F1 = 47.13) can degrade a stronger target setting (source Macro-F1 = 59.39) by up to 17.31 points. This demonstrates \textbf{performance-adaptive transferability}. 
(2) Compared with LLaMA-2, Qwen-3 exhibits much weaker transferability, yielding only limited gains under cross-dataset steering.

\subsubsection{On Pairs Combinations}
\label{sec: pc}

\begin{table}
\centering
\resizebox{\columnwidth}{!}{%
\begin{tabular}{@{}lllccc@{}}
\toprule
Backbone & Direction & Objective & Raw-FC & LIAR-RAW & AVeriTeC \\
\midrule
\multirow{8}{*}{LLaMA-2} & \multirow{4}{*}{Vertical} & w/o exp
& \begin{tabular}[c]{@{}l@{}}34.01\\(\textbf{↑7.57})\end{tabular}
& \begin{tabular}[c]{@{}l@{}}38.37\\(↑1.14)\end{tabular}
& \begin{tabular}[c]{@{}l@{}}74.86\\\underline{(↓1.05)}\end{tabular} \\
\cmidrule{3-6}
& & w/ exp
& \begin{tabular}[c]{@{}l@{}}62.17\\(↑1.58)\end{tabular}
& \begin{tabular}[c]{@{}l@{}}43.61\\(↑0.56)\end{tabular}
& \begin{tabular}[c]{@{}l@{}}85.86\\(↑1.24)\end{tabular} \\

\cmidrule{2-6}
& \multirow{4}{*}{Horizontal} & w/o exp
& \begin{tabular}[c]{@{}l@{}}34.82\\\textbf{(↑8.38)}\end{tabular}
& \begin{tabular}[c]{@{}l@{}}38.37\\(↑1.14)\end{tabular}
& \begin{tabular}[c]{@{}l@{}}74.79\\\underline{(↓1.12)}\end{tabular} \\
\cmidrule{3-6}
& & w/ exp
& \begin{tabular}[c]{@{}l@{}}62.64\\(↑2.05)\end{tabular}
& \begin{tabular}[c]{@{}l@{}}43.73\\(↑0.68)\end{tabular}
& \begin{tabular}[c]{@{}l@{}}82.92\\\underline{(↓1.70)}\end{tabular} \\

\midrule
\multirow{8}{*}{Qwen-3} & \multirow{4}{*}{Vertical} & w/o exp
& \begin{tabular}[c]{@{}l@{}}41.69\\(↑0.02)\end{tabular}
& \begin{tabular}[c]{@{}l@{}}41.91\\(↑0.19)\end{tabular}
& \begin{tabular}[c]{@{}l@{}}85.71\\(↑0.19)\end{tabular} \\
\cmidrule{3-6}
& & w/ exp
& \begin{tabular}[c]{@{}l@{}}58.88\\(↑0.53)\end{tabular}
& \begin{tabular}[c]{@{}l@{}}46.80\\(↑0.07)\end{tabular}
& \begin{tabular}[c]{@{}l@{}}88.62\\(↑0.60)\end{tabular} \\

\cmidrule{2-6}
& \multirow{4}{*}{Horizontal} & w/o exp
& \begin{tabular}[c]{@{}l@{}}42.10\\(↑0.43)\end{tabular}
& \begin{tabular}[c]{@{}l@{}}41.76\\(↑0.04)\end{tabular}
& \begin{tabular}[c]{@{}l@{}}85.89\\(↑0.37)\end{tabular} \\
\cmidrule{3-6}
& & w/ exp
& \begin{tabular}[c]{@{}l@{}}58.32\\(↓0.03)\end{tabular}
& \begin{tabular}[c]{@{}l@{}}47.04\\(↑0.31)\end{tabular}
& \begin{tabular}[c]{@{}l@{}}88.91\\(↑0.89)\end{tabular} \\

\bottomrule
\end{tabular}
}
\caption{Macro-F1 gains over SFT under different pairing strategies and training objectives.}
\vspace{-15pt}
\label{tab: pc}
\end{table}

To further test the flexibility of \textsc{REFLEX}, we construct inconsistent training pairs across objectives to induce more explicit disentanglement.

\textbf{Vertical steering} pairs BASE outputs without explanations ($c \rightarrow v_{\text{base}}$) and with explanation-augmented SFT outputs ($c \rightarrow v;exp_{\text{sft}}$). 
\textbf{Horizontal steering} pairs SFT outputs with and without explanations ($c \rightarrow v_{\text{sft}}$ vs.\ $c \rightarrow v;exp_{\text{sft}}$), independently of the standard \textsc{REFLEX} setup. For clarity, we omit evidence inputs for AVeriTeC hereafter.

As shown in Table~\ref{tab: pc}, both steering axes improve performance on all datasets except AVeriTeC, demonstrating the flexibility of the framework. Notably, explanation-guided vectors ($c \rightarrow v; exp$) can effectively steer explanation-free outputs ($c \rightarrow v$), yielding gains of up to 8.38 Macro-F1 points. This suggests that explanations can function as \textbf{internal activation signals} for reasoning. The slight drop ($>$ 0.1) on AVeriTeC is again likely due to saturation (Section~\ref{sec:obb}). Moreover, horizontal steering, which is derived solely from the SFT model, yields two drops on AVeriTeC, whereas vertical steering within \textsc{REFLEX} yields only one, further indicating the robustness of \textsc{REFLEX}.

\section{Analysis}

To better understand how \textsc{REFLEX} improves verdict prediction and explanation quality, we analyze explanation quality across all improved variants and conduct additional model-internal studies.

\subsection{Overall Evaluation}

As shown in Table~\ref{tab: eq-all} (Appendix~\ref{app: ae-on-exp}), automatic pointwise evaluation shows that steering generally shifts explanations toward lower misleadingness and higher informativeness, soundness, and readability. These trends suggest that \textsc{REFLEX} improves explanation faithfulness, especially in M and S, while also improving \textbf{stylistic plausibility}~\cite{agarwal2024faithfulness} across I, S, and R (see Appendix~\ref{app: con-ali}).

\begin{figure*}[htbp]
    \centering
    \begin{subfigure}{0.32\linewidth}
        \centering
        \includegraphics[width=\linewidth]{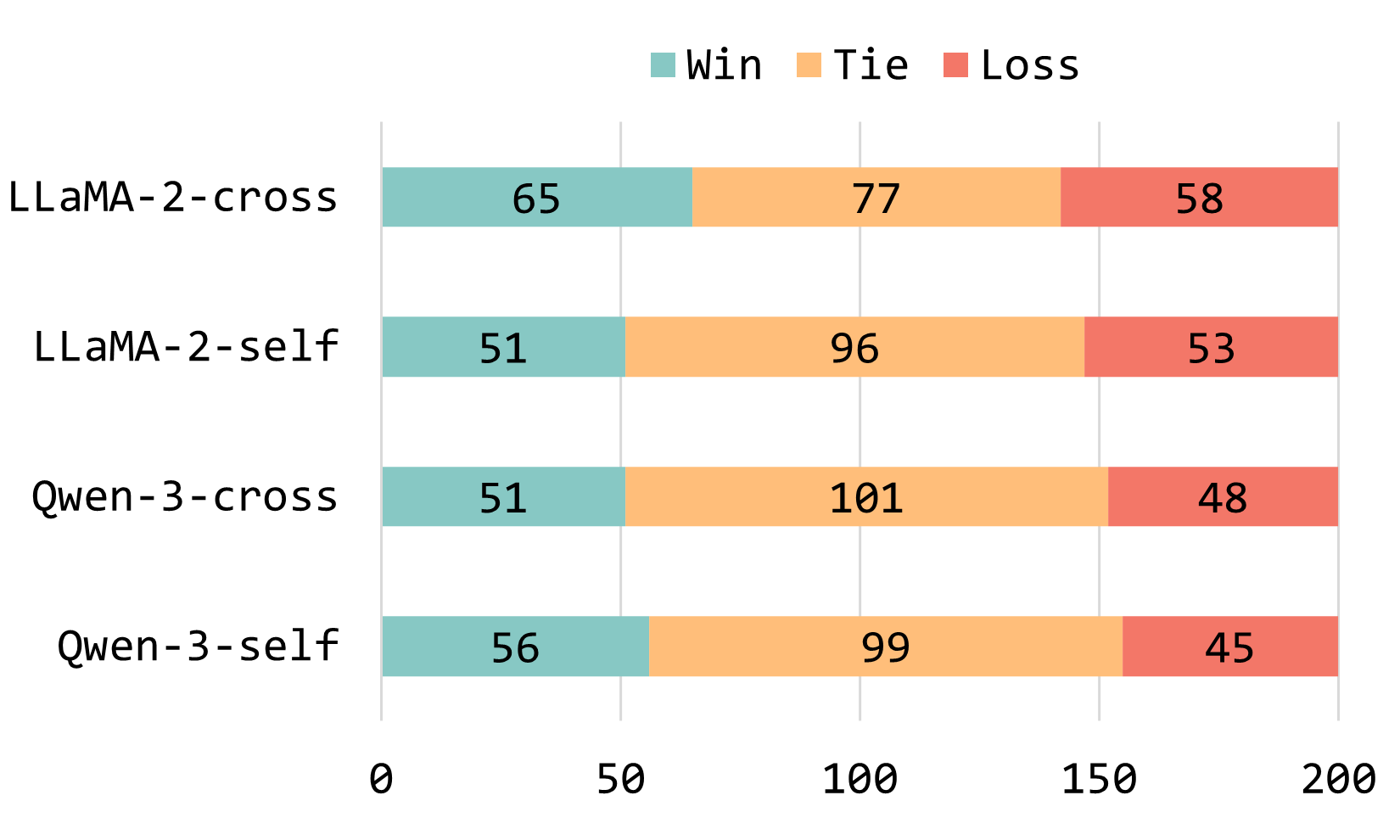}
        \caption{RAW-FC}
        \label{fig:rawfc-mi}
    \end{subfigure}
    \hfill
    \begin{subfigure}{0.32\linewidth}
        \centering
        \includegraphics[width=\linewidth]{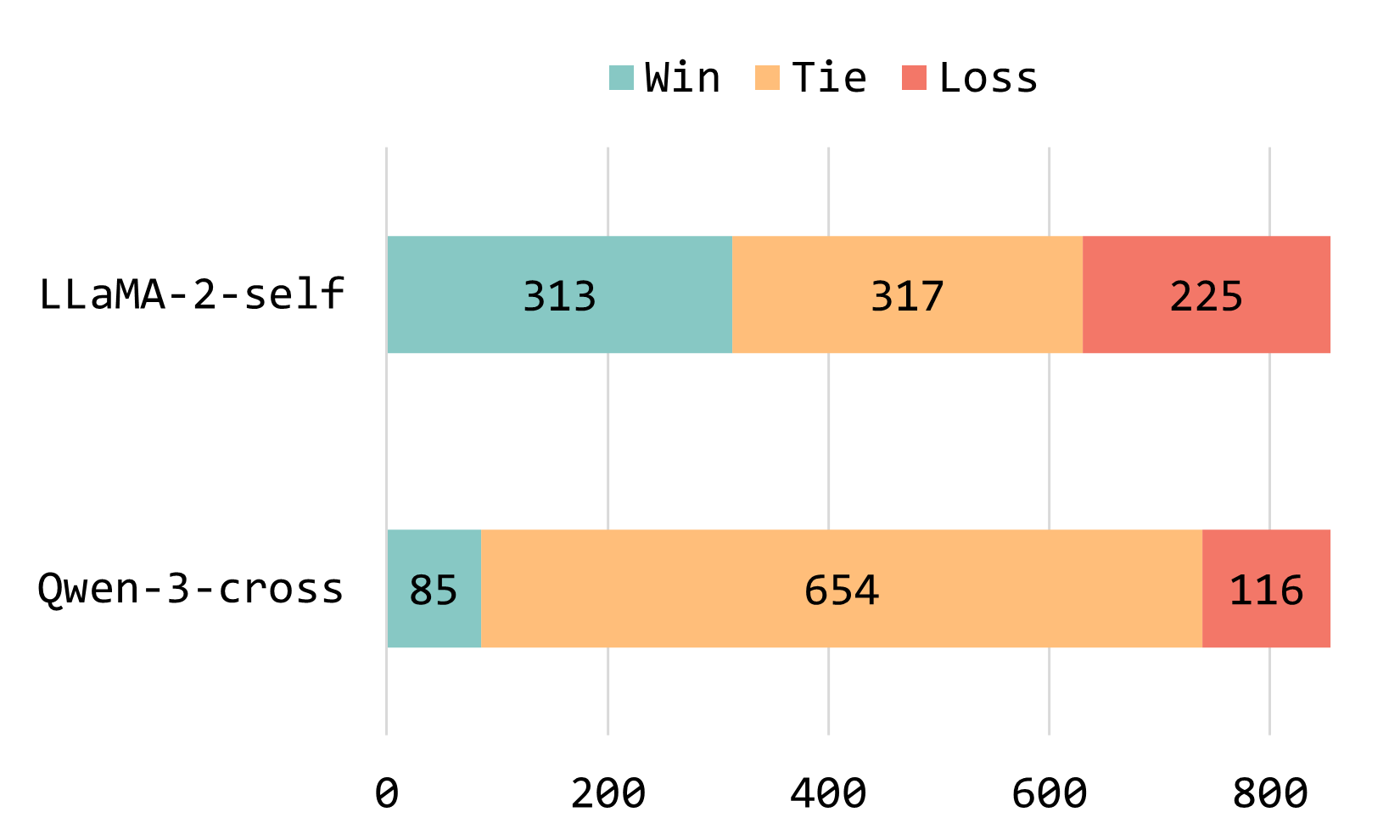}
        \caption{LIAR-RAW}
        \label{fig:liaraw-mis}
    \end{subfigure}
    \hfill
    \begin{subfigure}{0.32\linewidth}
        \centering
        \includegraphics[width=\linewidth]{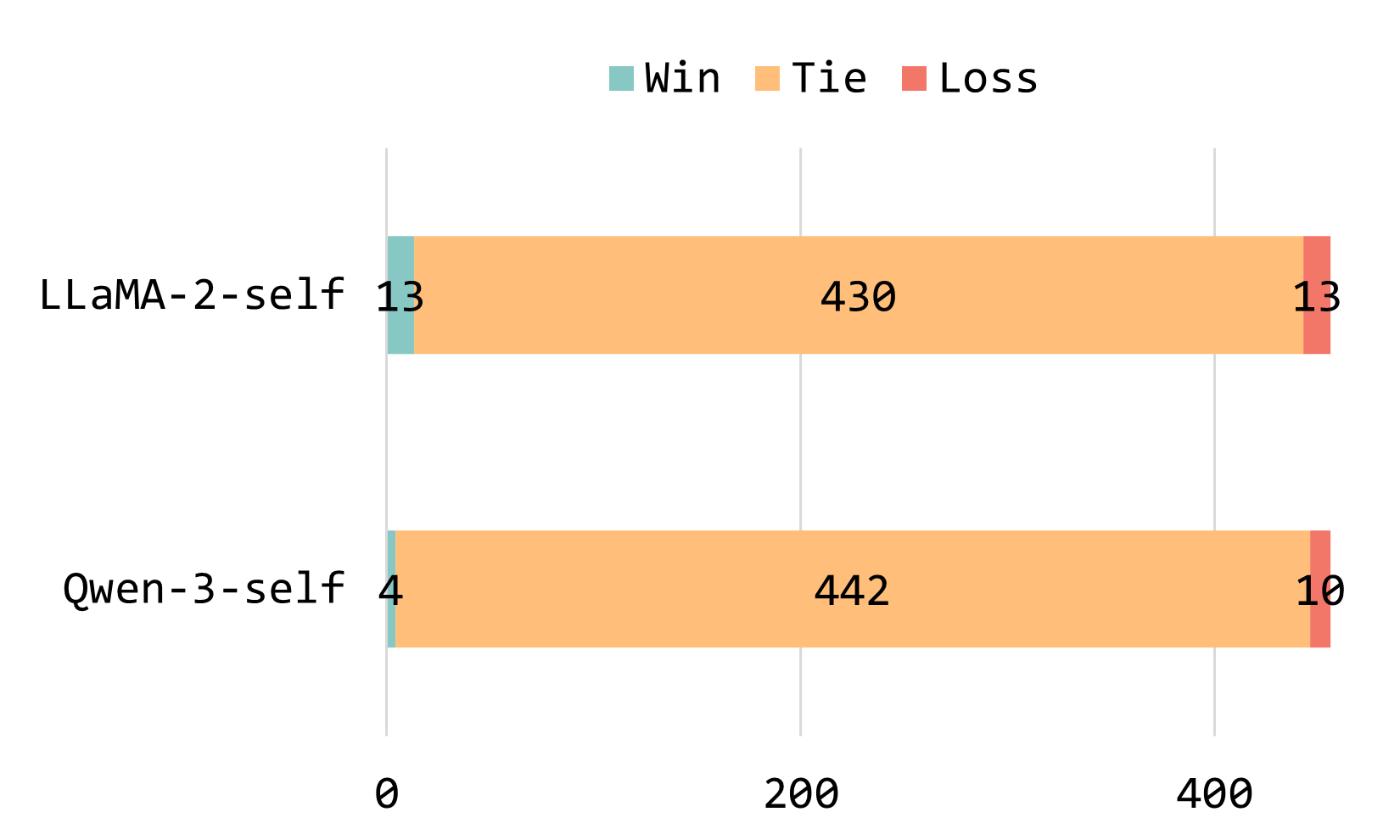}
        \caption{AVeriTeC}
        \label{sfig:averitec-mis}
    \end{subfigure}
    \vspace{5pt}
    \caption{Pairwise human evaluation of explanation non-misleadingness against the corresponding SFT models.}
    \label{fig:mis-dual-exp}
\end{figure*}

Because automatic evaluation may be unstable, we additionally conduct three rounds of pairwise human evaluation on the most factuality-relevant dimension, misleadingness, using Dual w/exp pairs. Figure~\ref{fig:mis-dual-exp} shows that \textsc{REFLEX} achieves higher or comparable non-misleadingness in most settings, except for Qwen-3, which is consistent with the automatic results. Overall error patterns are summarized in Appendix~\ref{app: error-ana}. In some cases, human results appear more favorable than automatic ones, possibly due to the relatively high standard error of automatic pointwise misleadingness scores (Appendix~\ref{app: ae-on-exp}).

\subsection{Disentanglement Effectiveness}
\label{sec: diver-ampl-eff}

\begin{figure}[t]
    \centering
    \begin{subfigure}{0.48\linewidth}
        \centering
        \includegraphics[width=\linewidth]{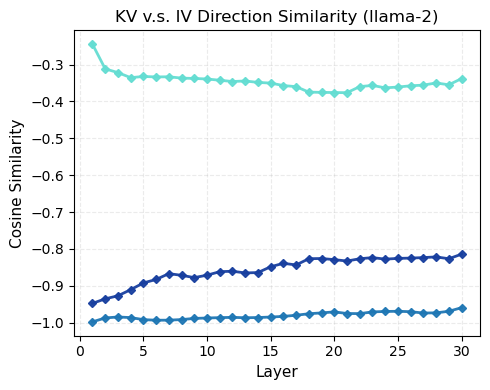}
        \caption{LLaMA-2}
        \label{fig:la2}
    \end{subfigure}
    \hfill
    \begin{subfigure}{0.48\linewidth}
        \centering
        \includegraphics[width=\linewidth]{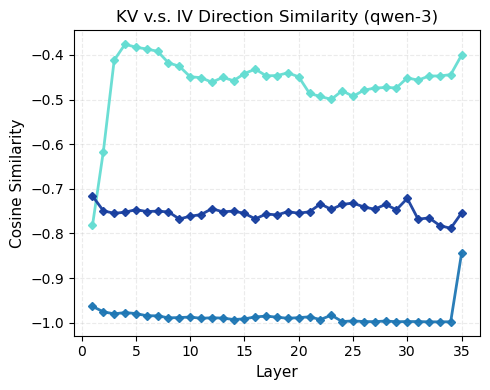}
        \caption{Qwen-3}
        \label{fig:ds-q3}
    \end{subfigure}
    \caption{Layer-wise cosine similarity between KVs and IVs for \textcolor{dualnoexp}{Dual w/o exp}, \textcolor{dualexp}{Dual w/exp}, and \textcolor{blue}{Single w/exp}.}
    \label{fig:direction}
    \vspace{-2pt}
\end{figure}

To assess the effectiveness of disentanglement, we analyze it from two perspectives: empirical patterns in explanation quality and structural patterns in the learned directions.

\begin{table}[]
\centering
\small
\begin{tabular}{@{}clccc@{}}
\toprule
\multirow{2}{*}{Vector} & \multirow{2}{*}{Split} & Factual Metric & \multicolumn{2}{c}{Stylistic Metrics} \\
\cmidrule(l){3-5}
 &  & M$\downarrow$ & I & S \\
\midrule
\multirow{3}{*}{KV} 
  & Total     & \textbf{1.58} & 4.56 & 4.72 \\
  & - Correct & \underline{1.52} & 4.56 & 4.74 \\
  & - Error   & \underline{1.74} & 4.54 & 4.67 \\
\midrule
\multirow{3}{*}{IV} 
  & Total     & 1.90 & \textbf{4.86} & \textbf{4.79} \\
  & - Correct & 1.89 & 4.85 & 4.79 \\
  & - Error   & 1.91 & 4.86 & 4.79 \\
\bottomrule
\end{tabular}
\caption{Average scores of explanations guided by the optimal KVs and IVs.}
\vspace{-15pt}
\label{tab: emp-v}
\end{table}

\paragraph{Empirical Validation.}
We first aggregate point-wise evaluation results by the type of optimal vector, excluding readability because it is affected by post-processing. We then divide the results into correct and erroneous verdict cases. Two patterns emerge.
\textbf{(1)} KVs are associated with lower misleadingness, suggesting a stronger relationship to \textbf{factual consistency}, whereas IVs are associated with higher informativeness and soundness, which is consistent with \textbf{stylistic improvement}. 
\textbf{(2)} Under KVs, the misleadingness gap between correct and erroneous cases is relatively large (1.52 vs.\ 1.74), which is consistent with the interpretation that KVs are more sensitive to factual differences. These findings support our distinction between IV-related reasoning style and KV-related factual sensitivity.

\paragraph{Representation Structure.} 
We next analyze the geometry of the learned directions. Specifically, we extract the optimal KV and IV at each layer for explanation-related pairs and compute their cosine similarity. Because similar patterns are observed across datasets (Appendix~\ref{app: full-ds}), we report the mean similarity.
Figure~\ref{fig:direction} shows that KV and IV directions tend to exhibit \textbf{negative similarity}, which is consistent with the hypothesis that they capture distinct aspects of the model’s internal behavior. We also observe a pattern that we refer to as the \textbf{divergence amplification effect}: settings with larger directional discrepancies between KVs and IVs also tend to show stronger empirical gains. Across backbones, Qwen-3 exhibits smaller KV-IV differences than LLaMA-2, which may help explain its weaker gains and lower transferability. Across pair constructions, Single w/exp exhibits the largest directional differences, which aligns with the stronger improvements observed when explanation-related pairs are used to steer lower-information settings without explanations (Section~\ref{sec: pc}). Additional analyses and case studies are provided in Appendix~\ref{app: more-disen-ana}.

\subsection{Statistical Correlations}
\label{sec: sc}

\begin{figure}[htbp!]
\centering
\includegraphics[width=0.75\linewidth]{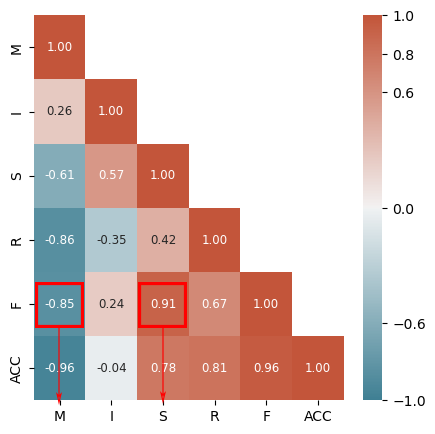} 
\caption{Correlation matrix between explanation quality and fact-checking performance.}
\label{fig:cm}
\end{figure}

To further examine whether improved explanations are associated with better verdict prediction, we analyze the correlation between verdict performance and point-wise explanation quality. Figure~\ref{fig:cm} shows that performance is most strongly associated with \textbf{faithfulness-related metrics}: both Macro-F1 and accuracy are strongly negatively correlated with misleadingness (-0.85 and -0.96), and strongly positively correlated with soundness (0.91 and 0.78). These results suggest that the gains of \textsc{REFLEX} are closely related to reduced faithfulness hallucination and improved logical coherence. Additional analyses and case studies are provided in Appendices~\ref{app: ana-cm} and~\ref{app: case-study}.

\section{Conclusion}

We introduced \textsc{REFLEX}, a simple and effective paradigm for explainable fact-checking that improves verdict prediction and explanation faithfulness by steering internal model activations rather than relying on external resources. 
Experiments across multiple settings demonstrate its effectiveness, generalizability, and flexibility.

\section*{Limitations}

\paragraph{Experiment Scale.}
Although we conducted extensive ablations across backbone models, cross-model transfer, pair combinations, and other settings, disk quota constraints and the overall experimental scope limited our study to LLaMA-2, Qwen-3, and Mistral-v0.1 (Appendix~\ref{app: error-ana}, Table~\ref{tab:segs_effect}). Moreover, research on real-world truthfulness does not necessarily follow standard scaling laws~\cite{li2023inference}. We therefore restrict our evaluation to models at the 7B--8B scale.

\paragraph{Label Scheme.}
Although our method improves both verdict accuracy and explanation quality, thereby strengthening their alignment, we adopt a three-way label scheme on LIAR-RAW, following prior work~\cite{wang2024explainable}. This choice reflects the difficulty LLMs face in generating explanations for overly fine-grained labels, such as ``pants-on-fire.'' Importantly, our goal is not merely to achieve state-of-the-art performance, but also to obtain interpretable insights.

\paragraph{Internal Design.}
Our internally guided paradigm reduces misinformation and response latency relative to methods that rely on external resources. Although internal knowledge may become outdated, this issue could in principle be mitigated through the lightweight activation-editing techniques on which we build (Section~\ref{sec: scifc}), enabling rapid adaptation to emerging fake news. However, existing time-shifted datasets such as VitaminC~\cite{schuster2021get} do not provide explanations, which prevents their direct use in our explainable setting.



\section*{Ethics Statement}
Although our experiments use open-source datasets and models, these models may still contain social biases inherited from their pretraining data. A systematic analysis or mitigation of such biases is beyond the scope of this work. Our primary focus is on reducing faithfulness hallucination and improving explanation quality.

\section*{Acknowledgments}
This project is supported by National Natural Science Foundation of China Young Scientists Fund (No. 62206233), RMGS (2025 First Processing Cycle), and the National Research Foundation, Singapore under its AI Singapore Programme (AISG Award No: AISG3-RP-2024-035). The experiment equipment (4xA100) is supported by Hong Kong Baptist University Strategic Development Fund.

We especially thank Yaxin Fan, Ziyang Luo, and Feng Jiang for their insightful discussions. Meanwhile, we thank all reviewers for their helpful comments.

\bibliography{custom}

\appendix
\newpage
\section*{\Large Appendix}

\section{Concept Alignment}
\label{app: con-ali}
\paragraph{Hallucination} typically refers to the generation of content that is either nonsensical or unfaithful to the provided source information~\citep{filippova2020controlled, maynez2020faithfulness}.

Its impact is application-dependent. In some domains, such as personalized role-play dialogue~\cite{kong2025sharp} and creative generation~\cite{hallu-creativity}, it can even be beneficial. However, in high-stakes societal domains such as law and fact-checking, as well as in knowledge-intensive tasks~\citep{sun2025causalabstain, sun2026facte}, hallucinations are harmful.

In the era of large language models (LLMs), recent work~\cite{hallu-llm} further categorizes hallucinations into two main types: factuality hallucinations and faithfulness hallucinations.

\textbf{\textit{Factuality hallucinations}} arise when LLMs produce outputs that are inconsistent with real-world facts or are potentially misleading.

\textbf{\textit{Faithfulness hallucinations}} concern whether generated content remains consistent with user instructions, provided context, and internal logical coherence. Accordingly, they can be categorized into instruction inconsistency, context inconsistency, and logical inconsistency. We argue that, in essence, this type of hallucination can be understood as \textbf{context inconsistency}.

\paragraph{Misinformation} refers to false or inaccurate information that is created deliberately and propagated either intentionally or unintentionally. In the context of social media, \citet{chen2024combating} adopts a broader view, treating misinformation as an umbrella term that encompasses all false or inaccurate information circulating on social media platforms. 

Under this definition, misinformation can be categorized into several types, including \textbf{fake news}, rumors, conspiracy theories, clickbait, \textbf{misleading claims}, and cherry-picking.

Moreover, they suggest that LLM-generated misinformation may exhibit \textbf{more deceptive styles}, thereby posing greater potential risk of real-world harm.

\paragraph{Misinformation vs.\ Hallucination.} \citet{chen2024combating} shows that hallucination is one source of misinformation, while \citet{hallu-llm} suggests that misinformation can also cause hallucination. Therefore, they can be regarded as intertwined phenomena that mutually influence each other in a bidirectional relationship.

\paragraph{Knowledge vs.\ Style.}
``IVs extracted from Q2 samples capture \textbf{style-dependent} veracity gains'' is supported by prior empirical work:

- OpenAI's fine-tuning guide notes that fine-tuning performs well in show-not-tell scenarios: ``show'' corresponds to stylistic patterns, whereas ``tell'' corresponds to knowledge injection.

- Several studies~\citep{ren2024learn,ghosal2024understanding,zhao2025style} argue that injecting specific or new knowledge via fine-tuning is comparatively difficult. \citet{zhao2025style} finds that fine-tuning on recent Wikipedia events (knowledge) performs worse, whereas fine-tuning on writing with a specific tone (style) yields near-perfect behavior.

- \citet{yan2025phd} shows that probability-based LLM inference in reasoning tasks often reflects memorized stylistic patterns.\\
\\
``KVs extracted from Q4 samples isolate \textbf{fact-sensitive} veracity drops'' is similarly supported by prior empirical findings.

- \citet{truth} points out that backbone models already encode extensive factual knowledge (Section 3.2.1, lines 203--207).

- \citet{gekhman2024does} reports that fine-tuning on new knowledge can increase hallucination risk, especially for dynamic knowledge in fact-checking.

- In alignment research, this phenomenon is sometimes informally referred to as an alignment tax~\cite{Leike_2022}.
\\

Transferred to our task, this literature suggests a natural mapping: \textbf{style} corresponds to \textbf{reasoning style}, while \textbf{knowledge} corresponds to \textbf{factual grounding}. This distinction motivates separating IV-driven style from KV-driven factual sensitivity.

\paragraph{Faithfulness vs. Plausibility on NLE.}
Natural Language Explanations (NLEs) refer to explanations expressed in natural language. According to~\cite{agarwal2024faithfulness}, recent work shows that modern LLMs can generate self-explanations (SEs), which aim to elicit intermediate reasoning steps to account for model outputs. They define the concepts of faithfulness and plausibility.

- \textbf{\textit{Plausibility}} concerns whether an explanation aligns with human reasoning and understanding. In this work, we treat it as a \textbf{stylistic pattern}.

- \textbf{\textit{Faithfulness}} evaluates whether an explanation accurately reflects the model’s underlying reasoning process. However, since we do not assume that LLMs possess genuine reasoning capabilities, our notion of faithfulness is framed through the lens of \textbf{faithfulness hallucinations}.

\section{Prompt Template}
\label{app: pt}
Following \citet{liu2022generated, chen2025towards}, the prompt template we use for training and inference is as follows: 
\begin{myquote}
A chat between a curious human and an artificial intelligence assistant. \\
You are a fact-checking assistant. \\ 
You are given a claim [(along with evidence sentences)].\\
Your task is to label the overall veracity of the claim based on your internal knowledge [or, on the provided evidence sentences].\\
Please reason step by step and explain how you reached your conclusion.\\
\\
The Label Definitions are as follows:\\
TRUE|true: The claim is verified as TRUE|true based on your knowledge.\\
FALSE|fase: The claim is verified as FALSE|false based on your knowledge.\\
HALF-TRUE|half: The claim is verified as half-true| due to insufficient knowledge leading to uncertainty, or because the claim itself is partially true.\\
\#\#\#\#\# or, for given evidence:

TRUE|true: The claim is verified as true or mostly-true by the evidence.\\
FALSE|false: The claim is verified as false or mostly false by the evidence.\\
HALF-TRUE|half: The claim is verified as half-true|half by the evidence, or the evidence can not prove the claim.\\ \#\#\#\#\# or, for AVeriTeC:

Supported: The claim is supported by the arguments and evidence presented.\\
Refuted: The claim is contradicted by the arguments and evidence presented.\\
Not Enough Evidence: There is not enough evidence to support or refute the claim.\\
\\
Based on the below claim and your own knowledge, determine the veracity of the claim. \\ 
\#\#\#\#\# or, for given evidence:\\
Based on the below claim and evidence, determine the veracity of the claim.\\
Please strictly output in the following format:\\
Verdict: [label].
Explanation: [your reasoning here]
\end{myquote}

Consistent with \citet{wang2024explainable}, the prompt template used to ask ChatGPT to conduct automatic evaluation is shown below:
\begin{myquote}
\label{app: em}
You are a helpful, harmless and precise assistant.\\
Please evaluate the quality of the explanations in prediction based on four metrics: misleadingness, informativeness, soundness, and readability, where 1 represented the poorest and 5 the best in addition to misleadingness. \\
The definitions of the metrics are:\\
(1) Misleadingness assesses whether the model’s explanation is consistent with the real veracity label of a claim, with a rating scale ranging from 1 (not misleading) to 5 (very misleading); \\
(2) Informativeness assesses whether the explanation provides new information, such as explaining the background and additional context, with a rating scale ranging from 1 (not informative) to 5 (very informative) \\
(3) Soundness describes whether the explanation seems valid and logical, with a rating scale ranging from 1 (not sound) to 5 (very sound)\\
(4) Readability evaluates whether the explanation follows proper grammar and structural rules, and whether the sentences in the explanation fit together and are easy to follow, with a rating scale ranging from 1 (poor) to 5 (excellent).\\
Please output scores in the following JSON format:\\
\verb|```|\\
\{\{\\
~~~~'Misleadingness': [an integer number from 1 to 5], \\
~~~~'Informativeness': [an integer number from 1 to 5], \\
~~~~'Soundness': [an integer number from 1 to 5], \\
~~~~'Readability': [an integer number from 1 to 5]\\
\}\}
\\
\verb|```|
\\
Claim: \{{claim\}} Label: \{{label\}}\\ Prediction: \{{verdict\}}\\ Score: 
\end{myquote} 

\section{Implementation Details}
\label{app: td}

For model training, Table~\ref{tab:hyper} shows the hyperparameter settings. Except for gradient accumulation steps and maximum context length, which depend on dataset size and sample length, we trained models on 1--4$\times$A100-80G GPUs with a learning rate of 2e-5, per-device batch size of 4, weight decay 0, warmup ratio 0.03, a cosine scheduler, bf16/tf32 precision, gradient checkpointing, and full-shard FSDP with auto wrapping. For LLaMA-2, we selected the epoch-2 checkpoint, which achieved the best balance between factual accuracy and instruction following. For Qwen-3, severe overfitting occurred at epoch 2 on some LIAR-RAW and RAW-FC variants (training accuracy exceeded test accuracy by 20--40\%), so we used the epoch-1 checkpoint instead.

For steering, consistent with the terminology of CAA~\cite{sv-llama2}, we use multipliers to denote steering magnitude. We sweep all layers and vary the multiplier over $\pm$0.5, $\pm$1, $\pm$1.5, and $\pm$2. Empirically, multipliers with magnitude $\geq 2$ (we additionally tested $\pm$2, $\pm$3, $\pm$5, $\pm$10, and $\pm$20) drive the model into non-instruction-following responses. The full optimal layers and multiplier hyperparameters across baseline trials and ablation studies are reported in Appendix~\ref{app: full-oh}.

We formalize disentangled steering as quadrant-conditioned activation probing over decoder hidden states in Algorithm~\ref{alg:disentangled_steering}.

Let the hidden activation at decoder layer $l$ and token position $t$ be:
\[
\mathbf{h}_{l,t} \in \mathbb{R}^d
\]
Two supervision datasets are constructed from cross-stage quadrants.
The Q2 dataset is defined as:
\[
\mathcal{D}_{IV} = \{ (\mathbf{h}^+_{l,i}, \mathbf{h}^-_{l,i}) \}_{i=1}^{N}
\]
where:
\begin{itemize}
    \item $\mathbf{h}^+$: SFT-correct reasoning activations
    \item $\mathbf{h}^-$: backbone-error activations
\end{itemize}
The Q4 dataset is defined as:
\[
\mathcal{D}_{KV} = \{ (\mathbf{h}^+_{l,i}, \mathbf{h}^-_{l,i}) \}_{i=1}^{N},
\]
where:
\begin{itemize}
    \item $\mathbf{h}^+$: backbone-correct activations
    \item $\mathbf{h}^-$: SFT-drift activations
\end{itemize}

\begin{algorithm}[htbp]
\caption{Disentangled Steering via Quadrant-Conditioned Probing}
\label{alg:disentangled_steering}
\KwIn{Q2 dataset (reasoning correction), Q4 dataset (factual preservation), decoder layers $L$, steering multipliers $\mathcal{A}$}
\KwOut{Optimal inference vector $IV^*$ and knowledge vector $KV^*$}

\ForEach{$l \in L$}{
    Extract Q2 activations $\{(h^{+}_{l,i}, h^{-}_{l,i})\}_{i=1}^{N}$\;
    Train logistic probe $p^{IV}_l(h)=\sigma(W^{IV\top}_l h + b^{IV}_l)$\;
    Normalize $IV_l = W^{IV}_l / \|W^{IV}_l\|$\;

    Extract Q4 activations $\{(h^{+}_{l,i}, h^{-}_{l,i})\}_{i=1}^{N}$\;
    Train logistic probe $p^{KV}_l(h)=\sigma(W^{KV\top}_l h + b^{KV}_l)$\;
    Normalize $KV_l = W^{KV}_l / \|W^{KV}_l\|$\;

    \ForEach{$\alpha \in \mathcal{A}$}{
        Apply IV steering: $h'_{l,t} = h_{l,t} + \alpha \, IV_l$\;
        Apply KV steering: $h'_{l,t} = h_{l,t} - \alpha \, KV_l$\;
        Compute improvement $\Delta P_{l,\alpha}$\;
    }

    Record $(l,\alpha_l^*) = \arg\max_{\alpha} \Delta P_{l,\alpha}$\;
}

Select $l^* = \arg\max_l \Delta P_{l,\alpha_l^*}$\;

Set:
$IV^* = \alpha_{l^*} IV_{l^*}$, \quad
$KV^* = \alpha_{l^*} KV_{l^*}$\;
\end{algorithm}

\section{Alignment between Probability and Performance}
\label{sec: alignment}

\begin{table}[t]
\centering
\caption{Maximum and minimum probability gaps ($\Delta$Prob.) between unsteered and steered outputs across 32 layers (step = 5), with corresponding Macro-F.}
\label{tab:alignment_layer_multiplier}
\resizebox{\columnwidth}{!}{%
\begin{tabular}{cccc}
\toprule
Layer & Multiplier & Macro-F (\%) & $\Delta$Prob.(\%) \\
\midrule
1  & 1.5  & \textbf{48.64} & \textbf{-4.32} \\
5  & 1.5  & 58.59 & 0.07  \\
10 & 1.5  & \textbf{64.99} & \textbf{0.14}  \\
10 & -1.5 & 59.61 & -0.15 \\
11 & -1.5 & 60.91 & -0.14 \\
15 & 1.5  & 63.09 & 0.11  \\
16 & -1.5 & 61.63 & -0.12 \\
17 & 1.5  & 59.88 & 0.11  \\
22 & -1.5 & 61.10 & -0.08 \\
22 & 1.5  & 60.57 & 0.07  \\
29 & 1.5  & 60.63 & 0.06  \\
30 & -1.5 & 61.13 & -0.06 \\
\midrule
\multicolumn{2}{l}{Correlation (Macro-F vs. Gap)} & \multicolumn{2}{c}{0.92} \\
\bottomrule
\end{tabular}}
\end{table}

Here, we report the probability gap and the corresponding Macro-F1 for Dual w/explanation pairs on RAW-FC with LLaMA-2.
As shown in Table~\ref{tab:alignment_layer_multiplier}, the overall correlation between probability shifts and verdict accuracy reaches 92\%, and their extrema are aligned: the largest probability gaps correspond to the largest performance shifts in both directions. 


\section{Dataset Details}
\label{app: data}

\begin{table}[t!]
\small
\centering
\caption{Summary statistics of dataset distributions. Label values 0-2 represent increasing veracity labels: \{False/Refuted, Half-True/Not Enough Evidence, True/Supported\}.}
\label{tab: dataset}
\begin{tabular}{@{}llrrrr@{}}
\toprule
Dataset & Split & 0 & 1 & 2 & Total \\
\midrule
\multirow{3}{*}{RAWFC} & train & 514 & 537 & 561 & 1,612 \\
      & eval  & 66  & 67  & 67  & 200   \\
      & test  & 66  & 67  & 67  & 200   \\
\midrule
\multirow{3}{*}{LIAR-RAW} & train & 2,568 & 1,336 & 2,264 & 6,168 \\
         & eval  & 410   & 159   & 292   & 861   \\
         & test  & 367   & 169   & 319   & 855   \\
\midrule
\multirow{3}{*}{AVeriTeC} & train & 1,742 & 849  & 282  & 2,873 \\
         & eval  & 305   & 35   & 122  & 462   \\
         & test  & 303   & 33   & 120  & 456   \\
\bottomrule
\end{tabular}
\end{table}

We use three fact-checking datasets: RAW-FC from Snopes\footnote{www.snopes.com}, LIAR-RAW~\citep{yang2022coarse} from PolitiFact\footnote{www.politifact.com}, and AVeriTeC. For RAW-FC and LIAR-RAW, evidence corresponds to the annotated relevant evidence (labeled as 1). Since AVeriTeC is natively multi-turn, we flatten each dialogue into a single-turn instance for consistency.

In terms of data structure, RAW-FC and LIAR-RAW follow the standard fact-checking format \{claim, evidence, label, explanation\}. In contrast, AVeriTeC decomposes fact-checking into a QA-style multi-turn verification process.

The datasets adopt different label schemes. RAW-FC uses \{True, Half-True, False\}, LIAR-RAW contains six labels, and AVeriTeC uses \{Supported, Not Enough Evidence, Conflicting Evidence/Cherrypicking, Refuted\}. To support joint verdict prediction and explanation generation in \textsc{REFLEX}, we unify labels as follows. For LIAR-RAW, we merge \{pants-fire, false, barely-true\} into False, retain Half-True, and merge \{mostly-true, true\} into True. For AVeriTeC, we discard Conflicting Evidence/Cherrypicking because of its ambiguity. We further remove LIAR-RAW instances without evidence and exclude AVeriTeC few-shot examples from validation to prevent data leakage.

As shown in Table~\ref{tab: dataset}, the training sample sizes of the three datasets are all below 10K, which matches the assumption of our paradigm, namely fine-tuning on limited data. Among them, RAW-FC is label-balanced, whereas LIAR-RAW and AVeriTeC are not. Since AVeriTeC does not release a test set to avoid leakage, we use its validation set for evaluation after removing training overlaps. Finally, because some baselines cannot handle dialogue-style inputs, baseline comparisons are reported only on RAW-FC and LIAR-RAW.

Since the human-annotated RAW-FC test set is relatively small, similar to other manually constructed benchmarks such as MT-Bench, we additionally compute confidence intervals using bootstrap resampling for the first ablation study (see Appendix~\ref{app: brrorf}).

\section{Automatic Evaluation Details}
\label{app: eval_details}

We conduct automatic pointwise evaluation on four dimensions: 
(1) \textbf{Misleadingness}, which assesses whether the model’s explanation is consistent with the true veracity label of a claim, with a rating scale from 1 (not misleading) to 5 (very misleading); 
(2) \textbf{Informativeness}, which assesses whether the explanation provides new information, such as background or additional context, with a rating scale from 1 (not informative) to 5 (very informative); 
(3) \textbf{Soundness}, which assesses whether the explanation appears valid and logical, with a rating scale from 1 (not sound) to 5 (very sound); 
(4) \textbf{Readability}, which assesses whether the explanation follows proper grammar and structure and is easy to follow, with a rating scale from 1 (poor) to 5 (excellent). 

Each dimension is rated on a five-point Likert scale, with higher scores indicating better quality except for misleadingness, which is inversely scored. The prompt template is shown in Appendix~\ref{app: em}. Following L-Defense, we use gpt-3.5-turbo-0613.

\section{Full Baselines Comparsions}
\label{app: compute-time}

Due to space constraints in the main text, we defer the full comparison of baselines and time complexity to this appendix.

As shown in Table~\ref{tab:time_cost}, distill/RAG-then-SFT paradigms (e.g., \textsc{FactLLaMA} and \textsc{L-Defense}) incur the highest time cost, because they require collecting multiple pieces of evidence per claim from closed-source models or search APIs, followed by task-specific fine-tuning. For $m$ claims and $n$ associated evidence items, both scale as $O(mn)$, with \textsc{L-Defense} further introducing adversarial explanations during distillation. Multi-agent approaches such as \textsc{RAV} remove training overhead but introduce substantial inference-time cost: each claim involves $Z$ agents, $(Z-1)$ interactions, and $Z$ inference steps, resulting in $O(mZ)$ complexity. RAG-based methods such as \textsc{HiSS} decompose each claim into $n$ sub-claims and iteratively perform retrieval, requiring repeated interactions with external APIs and yielding $O(mn)$ complexity.

In contrast, \textsc{REFLEX} eliminates reliance on closed-source APIs, external evidence, and multi-agent systems ($C=0, O=0, Z=0$). Although it requires training and vector extraction, once the optimal vectors are obtained they can be transferred to other models with only a constant number of inference steps per claim, resulting in linear time complexity $O(m)$.

We also include DeReC~\cite{qazi2025retrieval}, a recent fact-checking method. It is a retrieval-augmented fact verification framework that combines dense embedding-based evidence retrieval with FAISS-based similarity search and a DeBERTa-v3 classifier for veracity prediction. However, despite its efficiency and strong performance among claim-verification systems, it still underperforms \textsc{REFLEX} in Macro-F1. Following \citet{wang2024explainable}, and due to budget limitations, we exclude one additional model that also leverages external evidence~\cite{10.1162/tacl_a_00649}.

\begin{table*}[t]
\centering
\small
\setlength{\tabcolsep}{3pt} 
\begin{tabularx}{\textwidth}{
llcccc
>{\raggedright\arraybackslash}p{1.2cm}
*{4}{>{\centering\arraybackslash}p{1.2cm}}
}
\toprule
Paradigm & Method & No ED? & No AD? & Models & Macro-F & C & O  & I  & TC\\
\midrule

DI & - & $\checkmark$ & $\checkmark$ 
& LLaMA-2-7B-chat 
& 36.77 & $m$ & 0 & $m$ & $O(m)$ \\
DI+Retrieval & DeReC & $\times$  & $\checkmark$ &  \makecell{nomic-embed-text-v1.5 \\+ DeBERTa-v3-large}  &64.61& 0 & 1 & 1 & $O(l + \log n)$ \\

APIC & - & $\checkmark$ & $\times$ 
& ChatGPT 
& 44.43 & $m$ & 0 & $m$ & $O(m)$ \\

~ & - & $\times$ & $\times$ 
& ChatGPT 
& 39.31 & $m$ & 0 & $m$ & $O(m)$ \\

\midrule
\multirow{2}{*}{RAG} & HiSS & $\checkmark$ & $\times$ 
& ChatGPT 
& 53.90 & $m \times n$ & $2 \times m \times n$ & $m \times n$ & $O(mn)$ \\

~ & FactLLaMA & $\times$ & $\times$ 
& LLaMA-2-7B 
& 55.65 & $m \times n$ & 0 & $m$ & $O(mn)$ \\

\midrule
MAS & RAV & $\times$ & $\times$ 
& \makecell{LLaMA-3.1-70B-chat \\ $\times 3$} 
& 59.19 & $m \times z$ & $(z-1)\times m$ & $m \times z$ & $O(mZ)$ \\

\midrule
DTSFT & L-Defense & $\times$ & $\checkmark$ 
& \makecell{RoBERTa-large \\ + LLaMA-2-7B-chat} 
& 60.12 & $m \times n \times 2$ & 0 & $m \times 2$ & $O(mn)$ \\

~ & L-Defense & $\times$ & $\times$ 
& \makecell{RoBERTa-large \\ + ChatGPT} 
& 61.20 & $m \times n \times 2$ & 0 & $m \times 2$ & $O(mn)$ \\

SFT & FactLLaMA & $\times$ & $\checkmark$ 
& LLaMA-2-7B 
& 53.76 & $m$ & 0 & $m$ & $O(m)$ \\

\midrule
Ours & REFLEX & \textbf{$\checkmark$} & \textbf{$\checkmark$} 
& \textbf{LLaMA-2-7B}
& \textbf{64.99} & \textbf{0} & \textbf{0} & $m \times 2$ or \textbf{$m$}$^{\dagger}$ & \textbf{$O(m)$} \\

\bottomrule
\end{tabularx}
\caption{Full comparison of baselines in terms of paradigm, dependency, models involved, computational cost, and performance on RAW-FC. DI denotes Direct Inference. API denotes API Calls. MAS denotes Multi-Agent Systems. DTSFT denotes Distill-Then-SFT. ED denotes Evidence Dependency. AD denotes API Dependency. TC denotes Time Complexity. $C$: number of external API calls; $I$: number of inference steps; $O$: interaction overhead; $m$: number of claims; $n$: number of evidence items or sub-claims; $Z$: number of agents. $^{\dagger}$\textsc{REFLEX} requires a one-time post-training stage to extract transferable vectors. No retraining is needed when transferring to other models, so we also report the optimal time cost.}
\label{tab:time_cost}
\end{table*}


\noindent



\section{Explanation Length}
\label{app: el}
As Table~\ref{tab: exp-len-ba} shows, on RAW-FC, our outputs are shorter than those of L-Defense. On LIAR-RAW, they are shorter than all baselines, including the oracle, demonstrating that our paradigm produces concise and accurate explanations. Table~\ref{tab: exp-len-ab} further shows that EGS clearly reduces noisy patterns in model outputs across backbones and datasets.

\begin{table}[htbp!]
\centering
\vspace{-5pt}
\caption{Explanation lengths for our method and the baselines.}
\begin{tabular}{@{}lll@{}}
\toprule
Method & RAW-FC & LIAR-RAW \\ \midrule
Oracle & 201.68 & 220.75 \\
ChatGPT$_\text{w/evi}$ & 144.32 & 139.15 \\
ChatGPT$_\text{w/o evi}$ & \textbf{128.71} & 150.97 \\
L-Defense$_\text{ChatGPT}$ & 266.61 & 225.52 \\
L-Defense$_\text{LLaMA2}$& 305.50 & 175.38 \\
\midrule
\textbf{Ours} \\
S-EGS$_\text{LLaMA2}$ & 264.64 & \textbf{76.50} \\
w/o EGS & 787.55 & 118.06 \\ \bottomrule
\end{tabular}
\vspace{-10pt}
\label{tab: exp-len-ba}
\end{table}

\begin{table}[htbp!]
\caption{Hyperparameter settings for different models.}
\label{tab:hyper}
\centering
\resizebox{0.98\columnwidth}{!}{%
\begin{tabular}{@{}cccccc@{}}
\toprule
Dataset & \#train & x$\to$y & max-len & grad\_acc & epochs \\ 
\multicolumn{5}{c}{} & qwen / llama \\ 
\midrule
\multirow{3}{*}{RAW-FC} & \multirow{3}{*}{1,612} & c$\to$v & 512 & 1 & 1 / 2 \\
& & c;evi$\to$v & 1,024 & 1 & 2 / 2 \\
& & c$\to$v;exp & 4,096 & 1 & 2 / 2 \\
\midrule
\multirow{2}{*}{AVeriteC} & \multirow{2}{*}{2,873} & c;evi$\to$v & 512 & 4 & 2 / 2 \\
& & c;evi$\to$v;exp & 512 & 4 & 2 / 2 \\
\midrule
\multirow{3}{*}{LIAR-RAW} & \multirow{3}{*}{6,168} & c$\to$v & 256 & 8 & 1 / 2 \\
& & c;evi$\to$v & 512 & 8 & 1 / 2 \\
& & c$\to$v;exp & 1,024 & 8 & 2 / 2 \\
\bottomrule
\end{tabular}
}
\vspace{-10pt}
\end{table}

\begin{table}[htbp!]
\caption{Label distribution and few-shot ordering on LIAR-RAW. H denotes half-true, F false, and T true.}
\label{tab: label_distribution}
\centering
\resizebox{\columnwidth}{!}{%
\begin{tabular}{@{}cccccc@{}}
\toprule
Model & variant & order & k-shot & split & h/f/t \\ 
\midrule
\multirow{2}{*}{LLaMA-2} & \multirow{2}{*}{cross} & \multirow{2}{*}{ft\textbf{h}} & \multirow{2}{*}{3} & test  & \textbf{842}/8/5 \\
        &       &              &   & train & \textbf{6112}/37/19 \\
\midrule
\multirow{2}{*}{Qwen-3}  & \multirow{2}{*}{self}  & \multirow{2}{*}{fthth\textbf{f}} & \multirow{2}{*}{6} & test  & 189/\textbf{521}/145 \\
        &       &                &   & train & 1223/\textbf{3844}/1101 \\
\bottomrule
\end{tabular}
}
\vspace{6pt}
\end{table}

\begin{table}[htbp]
\vspace{-10pt}
\caption{Automatic and human evaluation of explanation quality. We report the best \textsc{REFLEX} variant on LLaMA-2.}
\centering
\resizebox{\linewidth}{!}{%
\begin{tabular}{l|cccc|cccc}
\hline
                  & \multicolumn{4}{c|}{\textbf{ChatGPT}} & \multicolumn{4}{c}{\textbf{Human}} \\ \cline{2-9} 
                  & \textbf{M$\downarrow $} & \textbf{I} & \textbf{S} & \textbf{R} & \textbf{M$\downarrow $} & \textbf{I} & \textbf{S} & \textbf{R} \\ \hline
Oracle            & 1.53 & 4.50 & 4.77& 4.77 & 1.47 & 3.61 & 3.89 & 3.86 \\ 
ChatGPT$_\text{c,v}$        & 2.07 & 4.43 & 4.67 & 4.73 & 2.22 & 3.22 & 3.38 & 3.57 \\
ChatGPT$_\text{c}$       & 2.33 & 4.17 & 4.43 & 4.63 & 2.68 & 2.68 & 2.84 & 3.27 \\ 
L-Defense$_\text{LLaMA2}$   & 1.87 & 4.50 & 4.67 & 4.67 & 2.12 & 3.48 & 3.37 & 3.49 \\
L-Defense$_\text{ChatGPT}$  & 1.77 & 4.40 & 4.60 & 4.53 & 1.97 & 3.68 & 3.52 & 3.56 \\ 

\midrule
\multicolumn{9}{l}{\textbf{Ours}} \\ 
S-EGS$_\text{LLaMA2}$     & \textbf{1.65} & \textbf{4.79} & \textbf{4.86} & \textbf{4.88} & \textbf{1.75} & \textbf{3.76} & \textbf{3.92} & \textbf{3.96} \\
w/o EGS          & 1.89 & 4.76 & 4.78 & 4.50 & 2.35 & 3.48 & 3.36 & 2.62 \\ \hline
\end{tabular}%
}

\vspace{-10pt}
\label{tab: me}
\end{table}

\begin{table*}[htbp!]
\vspace{-10pt}
\caption{Full statistics of hallucination ratio (HR), inference success rate (ISR), and data amounts across backbones and datasets. \textit{cross} denotes few-shot examples drawn from another dataset's training set, while \textit{self} denotes examples drawn from the model’s own validation set.}
\label{tab: full_statistics}
\centering
\begin{tabular}{@{}lllcccc@{}}
\toprule
\multirow{2}{*}{Backbone} & \multirow{2}{*}{Dataset} & \multirow{2}{*}{x->y} & \multicolumn{4}{c}{Statistics} \\ \cmidrule(l){4-7} 
& & & \#HC↓ & \#ISC↑ & HR↓ & ISR↑ \\ \midrule
\multirow{7}{*}{LLaMA-2} & \multirow{3}{*}{RAW-FC} & c-\textgreater v & 88 & 231 & 0.1236 &  \textbf{0.2567} \\
& & c-\textgreater v; exp$_{\text{self}}$ & 113 & 585 & 0.1768 & 0.6012 \\
& & c-\textgreater v; exp$_{\text{cross}}$ & 105 & 517 & 0.1502 & 0.5663 \\ 
\cmidrule(l){2-7} 
& \multirow{3}{*}{LIAR-RAW} & c-\textgreater v &  1,427 & 2,437 & 0.6731 & 0.6020 \\
& & c-\textgreater v; exp$_{\text{self}}$ & 895 & 2,207 & 0.3809 & 0.5781 \\
& & c-\textgreater v; exp$_{\text{cross}}$ & 1,304 & 3,600 &  \ \textbf{0.9546} & 0.7497 \\ 
\cmidrule(l){2-7} 
& AVeriTeC 
& c; evi -\textgreater v; exp$_{\text{self}}$ & 98 & 1,746 & 0.1033 & 0.9423 \\ \midrule
\multirow{7}{*}{Qwen-3} & \multirow{3}{*}{RAW-FC} & c-\textgreater v & 374 & 553 & 0.5351 & 0.6057 \\
& & c-\textgreater v; exp$_{\text{self}}$ & 161 & 502 & 0.2268 & 0.5565 \\ 
& & c-\textgreater v; exp$_{\text{cross}}$ & 156 & 482 & 0.2152 & 0.5434 \\ 
\cmidrule(l){2-7} 
& \multirow{3}{*}{LIAR-RAW} & c-\textgreater v &  903 & 1,893 & 0.3671 & 0.5105 \\
& & c-\textgreater v; exp$_{\text{self}}$ &  680 & 1,473 & 0.2388 & 0.4435 \\
& & c-\textgreater v; exp$_{\text{cross}}$ &  681 & 1,514 &  0.2426 & 0.4505 \\ 
\cmidrule(l){2-7} 
& AVeriTeC 
& c; evi -\textgreater v; exp$_{\text{self}}$ &  35 & 721 & \textbf{0.0174} & \textbf{0.9092} \\
\bottomrule
\end{tabular}
\label{tab: fss}
\end{table*}

\begin{table*}[htbp!]
\caption{The explanation length for ablation studies before and after S-EGS for improved variants. V denotes vertical steering, and H denotes horizontal steeering.}
\label{tab: exp-len-ab}
\centering
\begin{tabular}{@{}llccc@{}}
\toprule
Backbone   & Pair                 & RAW-FC & LIAR-RAW & AVeriTeC \\ 
\midrule
\multirow{5}{*}{LLaMA-2-7b} & Baseline             & 787.55 & 118.06 & 23.72 \\

           & Dual w/exp -- cross  & 278.42$\downarrow$ & --      & -- \\

           & Dual w/exp -- self   & 286.63$\downarrow$ & 67.29$\downarrow$  & 23.16$\downarrow$ \\

           & Single w/exp -- V    & 274.69$\downarrow$ & 76.82$\downarrow$  & 23.38$\downarrow$ \\

           & Single w/exp -- H    & 264.64$\downarrow$ & 76.50$\downarrow$  & -- \\ 
\midrule
\multirow{5}{*}{Qwen-3-7b}  & Baseline             & 997.04 & 290.03 & 24.37 \\

           & Dual w/exp -- cross  & 306.78$\downarrow$ & 84.88$\downarrow$  & -- \\

           & Dual w/exp -- self   & 293.92$\downarrow$ & --      & 24.41$\uparrow$ \\

           & Single w/exp -- V    & 310.00$\downarrow$ & 83.85$\downarrow$  & 24.23$\downarrow$ \\

           & Single w/exp -- H    & --      & 83.98$\downarrow$  & 22.89$\downarrow$ \\ 
\bottomrule
\end{tabular}
\vspace{-10pt}
\end{table*}

\begin{figure*}[htbp!]
\centering
\includegraphics[width=\linewidth, height=0.3\textheight, keepaspectratio]{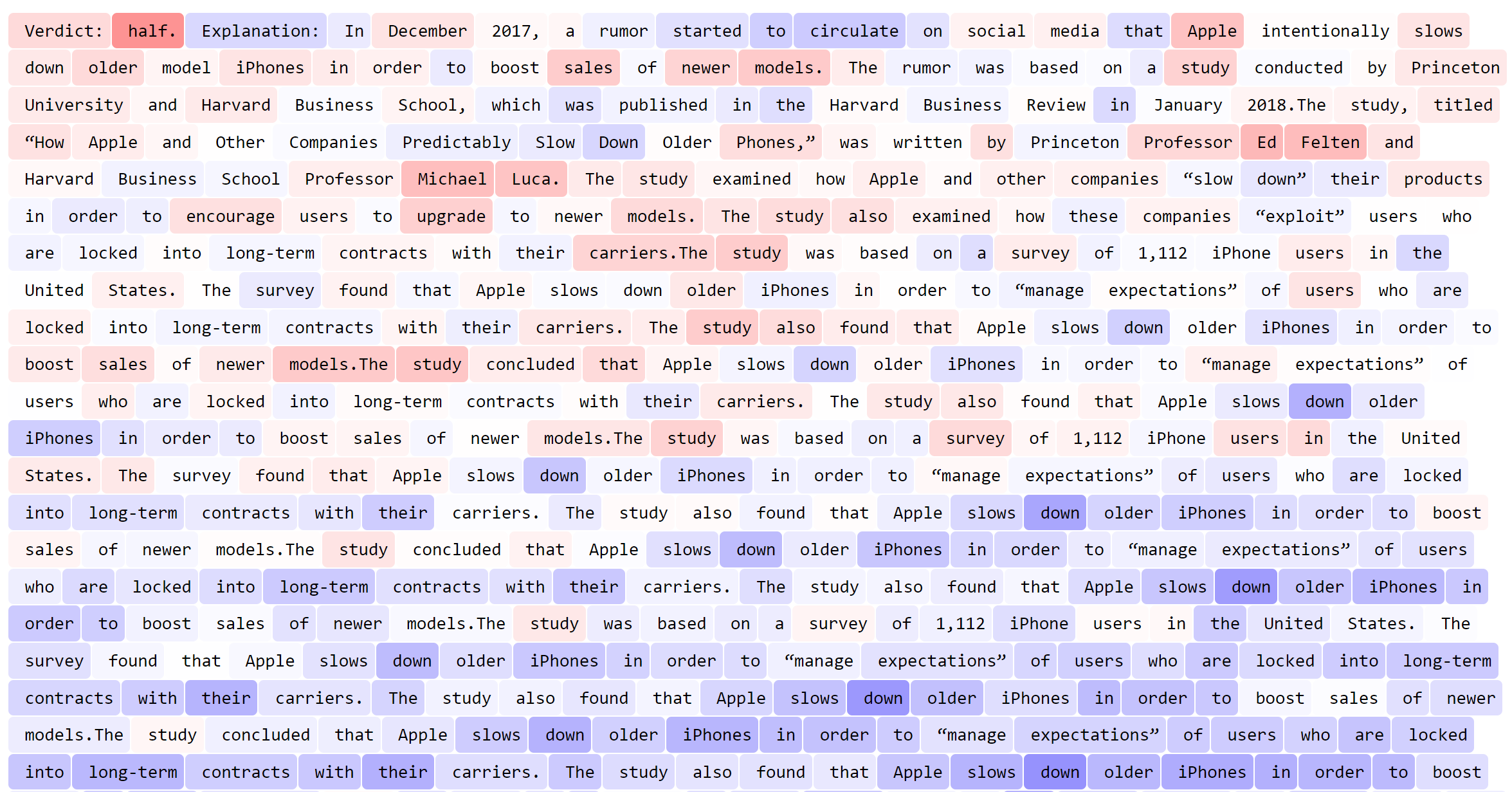} 
\caption{The redundancy noise pattern in LLaMA2 on RAW-FC, layer 10 with IV, multiplier 1.5. Red tokens denote alignment with optimal vector direction; blue denotes opposite.}
\label{fig:high-density}
\end{figure*}

\section{Human Evaluation Details}
\label{app: he}
All annotators were undergraduate students at a university where English is the official language. For the baseline experiments, we conducted pointwise evaluation of explanations on each dimension, following \citet{wang2024explainable}. Ten undergraduate annotators rated 30 randomly sampled test instances from RAW-FC using a 5-point Likert scale. Model identities were anonymized, and average scores were used as the final metric. Table~\ref{tab: me} shows that all four dimensions have correlation coefficients above 0.70 (0.95, 0.73, 0.85, 0.81), supporting the use of LLM-as-a-Judge~\cite{gu2025surveyllmasajudge}. Our S-EGS model outperforms the baselines on all four dimensions.

For inter-annotator agreement, because these are pointwise evaluations, we also computed Cronbach’s $\alpha$. The overall average $\alpha$ is 0.7896, which is acceptable by conventional standards.

For the ablation studies, we conducted pairwise evaluation focused on misleadingness, our main target dimension. Three undergraduate annotators rated each test sample from RAW-FC, LIAR-RAW, and AVeriTeC. Each annotator compared the explanations produced by two models and determined which one was more misleading. Model names remained anonymous, and the order of the two model outputs was randomly swapped. For inter-annotator agreement, we computed Fleiss’ Kappa. The overall average Kappa is 0.6963, indicating substantial agreement.

Notably, the case studies in Appendix~\ref{app: case-study} summarize common misinformation patterns and present annotator-verified examples. In these case studies, we additionally use ChatGPT-generated explanations for double validation.

\section{Label Distribution}
\label{app: ld}

As shown in Table~\ref{tab: label_distribution}, the two backbone models exhibit severe recency bias on LIAR-RAW, although this issue can be mitigated by prior methods~\citep{lu2022fantastically, min2022noisy, liu2022makes, zhangtempera, nguyen2023context}.

\section{Full Statistics on Backbone Ablations}
\label{app: hr}

Formally, we define the \textbf{hallucination rate (HR)} and \textbf{inference success rate (ISR)} as:
\begin{equation}
\text{HR} = \frac{\#\text{error after SFT}}{\#\text{correct on BASE}},
\end{equation}
\begin{equation}
\text{ISR} = \frac{\#\text{correct after SFT}}{\#\text{error on BASE}}.
\end{equation}
We compute HR and ISR across models and datasets. The full statistics, including self-refined sample sizes, are shown in Table~\ref{tab: full_statistics}.

\section{Bootstrap Resampling Results on RAW-FC}
\label{app: brrorf}
As shown in Table~\ref{tab: stat-rawfc-results}, we find that: 
(1) the bootstrap Macro-F1 mean closely matches the original Macro-F1 scores across all settings; 
(2) performance gains remain consistent under resampling; 
(3) the 95\% confidence intervals are stable. 
Together, these results indicate that our findings are statistically stable.

\begin{table*}[t]
\centering
\small
\begin{tabular}{llllcccc}
\toprule
Backbone & Stage & $x \rightarrow y$ & Macro-F1 & $\Delta$Gain & Bootstrap FS Mean & Bootstrap $\Delta$Gain & 95\% CI Width/2 \\
\midrule

\multirow{10}{*}{LLaMA-2} & \multirow{4}{*}{BASE} & $c \rightarrow v$ & 35.61 & — & 35.46 & — & $\pm 5.31$ \\
~ & ~ & $c; evi \rightarrow v$ & 27.08 & — & 28.08 & — & $\pm 6.79$ \\
~ & ~ & $c \rightarrow v; exp_{cross}$ & 34.41 & — & 34.16 & — & $\pm 6.42$ \\
~ & ~ & $c \rightarrow v; exp_{self}$ & 31.68 & — & 31.54 & — & $\pm 5.14$ \\
\cmidrule{2-8}
~ & \multirow{3}{*}{SFT} & $c \rightarrow v$ & 26.44 & -9.17 & 26.31 & -9.15 & $\pm 5.73$ \\
~ & ~ & $c; evi \rightarrow v$ & 44.85 & 17.77 & 45.01 & 16.93 & $\pm 6.75$ \\
~ & ~ & $c \rightarrow v; exp$ & 60.59 & 26.18 & 60.42 & 26.26 & $\pm 6.62$ \\
\cmidrule{2-8}
~ & \multirow{3}{*}{S-EGS} & $c \rightarrow v$ & 31.47 & 5.03 & 31.25 & 4.94 & $\pm 6.82$ \\
~ & ~ & $c \rightarrow v; exp_{cross}$ & 64.99 & 4.40 & 64.84 & 4.42 & $\pm 6.57$ \\
~ & ~ & $c \rightarrow v; exp_{self}$ & 61.81 & 1.22 & 61.64 & 1.22 & $\pm 6.60$ \\

\midrule

\multirow{10}{*}{Qwen-3} & \multirow{4}{*}{BASE} & $c \rightarrow v$ & 46.54 & — & 46.36 & — & $\pm 7.03$ \\
~ & ~ & $c; evi \rightarrow v$ & 46.23 & — & 46.00 & — & $\pm 7.09$ \\
~ & ~ & $c \rightarrow v; exp_{cross}$ & 46.66 & — & 46.53 & — & $\pm 6.87$ \\
~ & ~ & $c \rightarrow v; exp_{self}$ & 48.86 & — & 48.72 & — & $\pm 6.95$ \\
\cmidrule{2-8}
~ & \multirow{3}{*}{SFT} & $c \rightarrow v$ & 41.67 & -4.87 & 41.46 & -4.90 & $\pm 6.74$ \\
~ & ~ & $c; evi \rightarrow v$ & 63.17 & 16.94 & 62.99 & 16.99 & $\pm 6.56$ \\
~ & ~ & $c \rightarrow v; exp$ & 58.35 & 9.49 & 58.19 & 9.47 & $\pm 6.85$ \\
\cmidrule{2-8}
~ & \multirow{3}{*}{S-EGS} & $c \rightarrow v$ & 41.69 & 0.02 & 41.52 & 0.06 & $\pm 6.81$ \\
~ & ~ & $c \rightarrow v; exp_{cross}$ & 59.39 & 1.04 & 59.23 & 1.04 & $\pm 6.80$ \\
~ & ~ & $c \rightarrow v; exp_{self}$ & 58.86 & 0.51 & 58.70 & 0.51 & $\pm 6.80$ \\

\bottomrule
\end{tabular}
\caption{Bootstrap resampling results for the first ablation study on RAW-FC.}
\label{tab: stat-rawfc-results}
\end{table*}

\section{Full Cross-Model Transfer Results}
\label{app: full-cmtr}

The full cross-model transfer results are shown in Table~\ref{tab:cross_transfer}. Here, $\Delta$ denotes the difference relative to the corresponding SFT model.

\begin{table*}[htbp]
\centering
\caption{Full cross-model transfer results in Macro-F1. Corr$\Delta$=0.78 and Corr(Source Macro-F, Target $\Delta$)=0.88. Corr.\ denotes the Pearson correlation coefficient. V denotes vertical steering, and H denotes horizontal steering.}
\label{tab:cross_transfer}
\resizebox{\textwidth}{!}{%
\begin{tabular}{lllcclcc}
\toprule
Model & Source & Variant & Macro-F & $\Delta$ & Target & Macro-F & $\Delta$ \\
\midrule
\multirow{6}{*}{LLaMA-2} 
& \multirow{3}{*}{RAW-FC} 
& Dual w/ exp.           & 64.99 & +4.40 & \multirow{3}{*}{Liar-RAW} & 50.59 & +7.54 \\
&                        & Single w/ exp. - V & 62.17 & +1.58 &                         & 49.81 & +6.76 \\
&                        & Single w/ exp. - H & 62.64 & +2.05 &                         & 50.11 & +7.06 \\
\cmidrule(lr){2-8}
& \multirow{3}{*}{Liar-RAW} 
& Dual w/ exp.           & 43.06 & +0.01 & \multirow{3}{*}{RAW-FC}   & 47.20 & -13.39 \\
&                        & Single w/ exp. - V & 43.61 & +0.56 &                         & 51.77 & -8.82 \\
&                        & Single w/ exp. - H & 43.73 & +0.68 &                         & 52.72 & -7.87 \\
\midrule
\multirow{5}{*}{Qwen-3} 
& \multirow{2}{*}{RAW-FC} 
& Dual w/ exp.           & 59.39 & +1.04 & \multirow{2}{*}{Liar-RAW} & 45.12 & -1.61 \\
&                        & Single w/ exp. - V & 58.88 & +0.53 &                         & 45.28 & -1.45 \\
\cmidrule(lr){2-8}
& \multirow{3}{*}{Liar-RAW} 
& Dual w/ exp.           & 47.13 & +0.40 & \multirow{3}{*}{RAW-FC}   & 41.04 & -17.31 \\
&                        & Single w/ exp. - V & 46.80 & +0.07 &                         & 40.40 & -17.95 \\
&                        & Single w/ exp. - H & 47.04 & +0.31 &                         & 40.40 & -17.95 \\
\bottomrule
\end{tabular}
}
\end{table*}

\section{Full Direction Simialarity of Vectors}
\label{app: full-ds}

\begin{figure*}[t]
    \vspace{20pt}
    \centering
    \begin{subfigure}{0.48\linewidth}
        \centering
        \includegraphics[width=\linewidth]{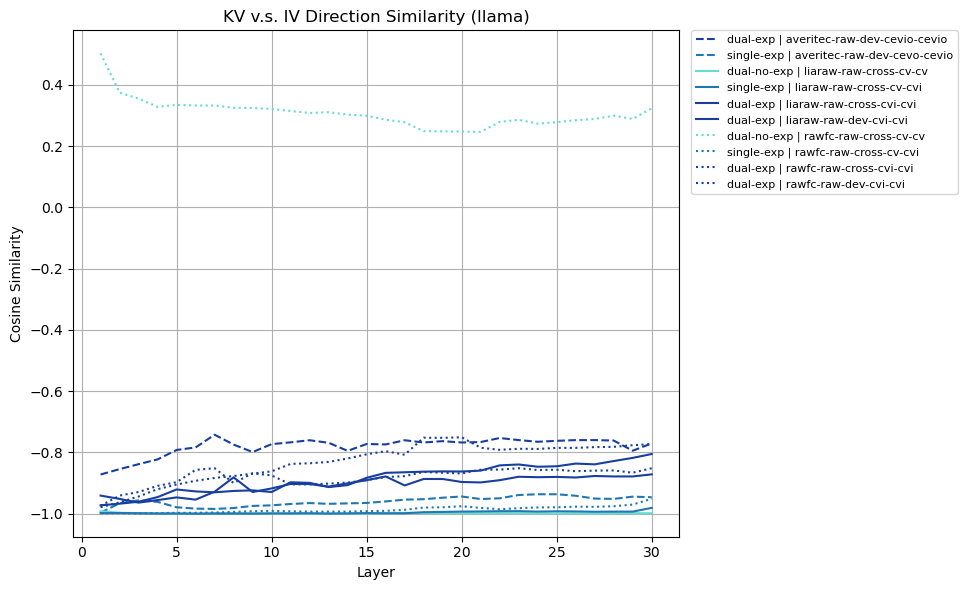}
        \caption{LLaMA-2}
        \label{fig:la2}
    \end{subfigure}
    \hfill
    \begin{subfigure}{0.48\linewidth}
        \centering
        \includegraphics[width=\linewidth]{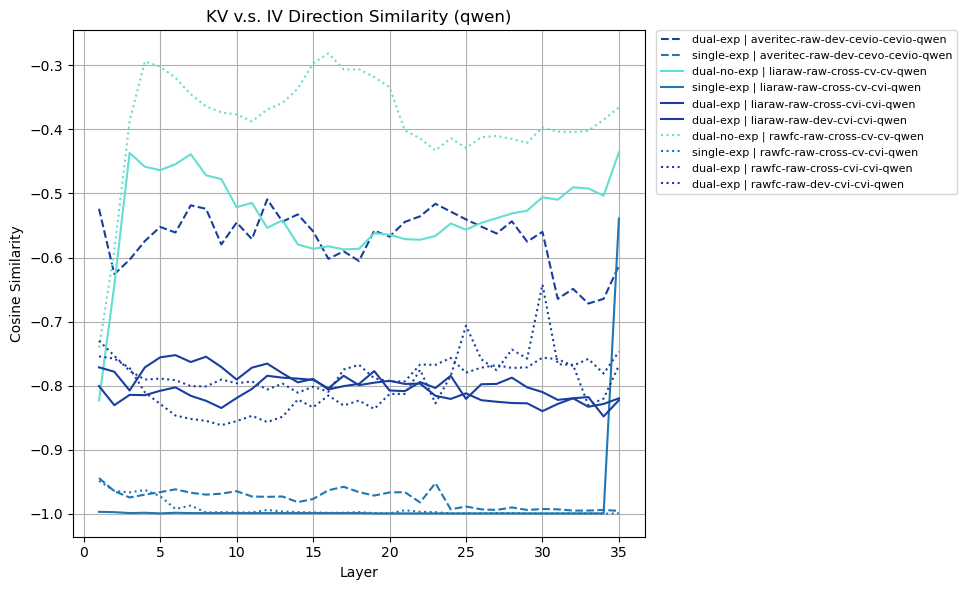}
        \caption{Qwen-3}
        \label{fig:ds-q3}
    \end{subfigure}
    \caption{Full direction similarity of KV and IV at each layer based on cosine similarity. Colors denote \textcolor{dualnoexp}{Dual w/o exp}, \textcolor{dualexp}{Dual w/exp}, and \textcolor{blue}{Single w/exp}. The solid line denotes LIAR-RAW, the dashed line AVeriTeC, and the dotted line RAW-FC.}
    \label{fig: full-direction}
    \vspace{-10pt}
\end{figure*}

The full direction similarity, without averaging across datasets, is shown in Figure~\ref{fig: full-direction}.

\section{Optimal Hyperparameters Results}
\label{app: full-oh}
The full optimal hyperparameters, including layers and multipliers across baseline and ablation settings, are shown in Tables~\ref{tab:hyper_la2} and~\ref{tab:hyper_q3}. Because horizontal steering is not specific to \textsc{REFLEX}, we do not report its direction.

As shown in both tables, most KV multipliers are negative, while most IV multipliers are positive, indicating KV suppression and IV amplification, respectively. For the bolded cases, the majority of the corresponding probabilities are negative, which further supports KV suppression and IV amplification. The only exceptions are the two underlined KVs; accordingly, they yield only 0.01 and 0.19 Macro-F1-point gains. 

We also observe that the optimal number of vectors required for IV is larger than that required for KV. This is because suppressing erroneous directions in KV is a \emph{necessary condition} for ensuring factual consistency between the verdict and the explanation, whereas amplifying correct directions in IV serves as a \emph{sufficient condition} that directly aligns the reasoning process in explanations with successful verdict prediction. In general, sufficient conditions are less restrictive than necessary ones.

\begin{table}[htbp]
\centering
\caption{Optimal steering hyperparameters across datasets on LLaMA-2.}
\label{tab:hyper_la2}
\small
\resizebox{\linewidth}{!}{%
\begin{tabular}{lccccc}
\toprule
\multirow[c]{2}{*}{Dataset} 
 & \multicolumn{5}{c}{\textbf{LLaMA-2}} \\
\cmidrule(lr){2-6}
 & Pairs & Target & Vector & Layer & Multiplier \\
\midrule
\multirow{7}{*}{RAW-FC}
 & Dual w/exp-cross   & w/exp   & IV & 10 & 1.5 \\
 \cmidrule{2-6}
 & Dual w/exp-self    & w/exp   & IV & 13 & 1.5 \\
  \cmidrule{2-6}
 & \multirow{2}{*}{Single w/exp - V}
                       & w/exp   & IV & 2  & 1   \\
 &                     & w/o exp & IV & 1  & 1.5 \\
  \cmidrule{2-6}
 & Dual w/o exp       & w/o exp & IV & 1  & 1.5 \\
  \cmidrule{2-6}
 & \multirow{2}{*}{Single w/exp - H}
                       & w/exp   & -- & 2  & -1  \\
 &                     & w/o exp & -- & 1  & 1.5 \\
\midrule
\multirow{6}{*}{LIAR-RAW}
 & \multirow{2}{*}{Dual w/exp-self}
                       & \underline{\textbf{w/exp}}   & KV & 1 & \textbf{1.5} \\
 &                     & \textbf{w/o exp} & KV & 1 & \textbf{1}   \\
  \cmidrule{2-6}
 & \multirow{2}{*}{Single w/exp - V}
                       & w/exp   & IV & 1 & 1   \\
 &                     & w/o exp & IV & 2 & 1   \\
  \cmidrule{2-6}
 & \multirow{2}{*}{Single w/exp - H}
                       & w/exp   & -- & 1 & 1   \\
 &                     & w/o exp & -- & 2 & -1  \\
\midrule
\multirow{2}{*}{AVeriTeC}
 & Dual w/exp-self    & w/exp & KV & 11 & -1.5 \\
 & Single w/exp - V   & w/exp & IV & 1  & 1    \\
\bottomrule
\end{tabular}}
\end{table}

\begin{table}[htbp]
\centering
\caption{Optimal steering hyperparameters across datasets on Qwen-3.}
\label{tab:hyper_q3}
\small
\resizebox{\linewidth}{!}{%
\begin{tabular}{lccccc}
\toprule
\multirow[c]{2}{*}{Dataset} 
 & \multicolumn{5}{c}{\textbf{Qwen3}} \\
\cmidrule(lr){2-6}
 & Pairs & Target & Vector & Layer & Multiplier \\
\midrule
\multirow{6}{*}{RAW-FC}
 & Dual w/exp-cross   & w/exp   & IV & 10 & 1.5 \\
  \cmidrule{2-6}
 & Dual w/exp-self    & w/exp   & IV & 10 & 1.5 \\
  \cmidrule{2-6}
 & \multirow{2}{*}{Single w/exp - V}   & w/exp   & IV & 16 & 1.5 \\
 &                     & w/o exp & KV & 13 & -1.5 \\
  \cmidrule{2-6}
 & Dual w/o exp       & \textbf{w/o exp} & IV & 17 & \textbf{-1.5} \\
  \cmidrule{2-6}
 & Single w/exp - H   & w/o exp & -- & 13 & -1.5 \\
\midrule
\multirow{6}{*}{LIAR-RAW}
 & Dual w/exp-cross   & w/exp   & KV & 15 & -1.5 \\
 \cmidrule{2-6}
 & Dual w/o exp       & w/o exp & KV & 2  & -1   \\
 \cmidrule{2-6}
 & \multirow{2}{*}{Single w/exp - V}   & w/exp   & IV & 4  & 1.5  \\
 &                     & w/o exp & IV & 6  & 1.5  \\
 \cmidrule{2-6}
 & \multirow{2}{*}{Single w/exp - H}   & w/exp   & -- & 1  & 1.5  \\
 &                     & w/o exp & -- & 6  & 1.5  \\
\midrule
\multirow{5}{*}{AVeriTeC}
 & Dual w/exp-self    & \underline{\textbf{w/exp}}   & KV & 10 & \textbf{1.5}  \\
 \cmidrule{2-6}
 & \multirow{2}{*}{Single w/exp - V}   & \textbf{w/exp}   & IV & 7  & \textbf{-1.5} \\
 &                     & \textbf{w/o exp} & IV & 3  & \textbf{-1.5} \\
 \cmidrule{2-6}
 & \multirow{2}{*}{Single w/exp - H}   & w/exp   & -- & 1  & 1.5 \\
 &                     & w/o exp & -- & 3  & 1.5  \\
\bottomrule
\end{tabular}}
\end{table}

\section{Total Error Analysis}
\label{app: error-ana}
Overall, across 38 experiments (excluding performance-adaptive transfer trials), the Macro-F1 score drops by more than 0.01 only six times: -0.28, -0.20, -0.03, -1.05, -1.12, and -1.17.
Detailed analyses are provided below:

(1) The decreases of -0.28 and -0.20 occur only on LIAR-RAW. As noted in Section~\ref{sec:obb} and Appendix~\ref{app: ld} (Table~\ref{tab: label_distribution}), this is due to severe \textbf{recency bias} in few-shot learning. Although prior methods~\citep{lu2022fantastically,min2022noisy,liu2022makes,zhangtempera,nguyen2023context} can mitigate this bias, we deliberately avoid them to maintain experimental consistency.

(2) For the larger decreases (-1.05 to -1.17): beyond the explanation in Section~\ref{sec:obb}, the \textbf{AVeriTeC} dataset decomposes reasoning into a dialogue format that closely matches our Stage-1 training pipeline. This effectively represents an upper bound of our optimization setting. In addition, the decoupled AVeriTeC test set is inherently simpler than the other datasets; even SFT outperforms the others by roughly 20--40\% according to Table~\ref{tab: bb}. Strictly speaking, \textbf{horizontal steering} vectors are extracted only from SFT models and are not specific to \textsc{REFLEX}. Consequently, horizontal steering fails twice on AVeriTeC, whereas vertical steering fails only once, which further illustrates the robustness of \textsc{REFLEX}.

(3) Smaller gains on \textbf{Qwen-3} can be explained as follows. First, as discussed in Section~\ref{sec: diver-ampl-eff}, we observe a divergence amplification effect. Except for the Dual w/o exp setting, which contains limited information and therefore shows only minor backbone differences, Qwen-3 exhibits smaller divergences than LLaMA-2 in Dual w/exp and Single w/exp settings, which may explain its reduced gains and weaker transferability. Second, as noted in Appendix~\ref{app: td}, Qwen-3 shows severe overfitting, with training Macro-F1 exceeding test performance by 20--40\%. Although we selected checkpoints to reduce overfitting, a gap of roughly 10\% remains. Overfitting in the Qwen-3 series is also commonly reported on reasoning tasks.

To further demonstrate generalizability, inspired by one reviewer’s observation on vertical and horizontal steering, we also ran experiments on Mistral-v0.1 under the w/exp setting on RAW-FC. As shown in Table~\ref{tab:segs_effect}, \textsc{REFLEX} still achieves an 8.88-point gain under vertical steering.

\begin{table}[t]
\centering
\small
\begin{tabular}{llccc}
\toprule
Backbone & Direction & SFT & S-EGS & $\Delta$Gain \\
\midrule

\multirow{2}{*}{Mistral-v0.1} & V  & 42.03 & 50.91 & 8.88 \\
~             & H & 42.03 & 50.36 & 8.33 \\

\bottomrule
\end{tabular}
\caption{Macro-F1 scores under different steering directions on the Mistral-v0.1 backbone. V denotes vertical steering and H denotes horizontal steering.}
\label{tab:segs_effect}
\end{table}

\section{More Analysis of Disentanglement Effectiveness}
\label{app: more-disen-ana}
To further support disentanglement beyond the main results on empirical validation and representation structure, we provide additional analyses from complementary perspectives, including model directions, optimal layers, and case studies.

\begin{table}[t]
\caption{Macro-F1 scores for model-direction experiments. Red indicates effective settings, and blue indicates ineffective settings.}
\centering
\resizebox{\columnwidth}{!}{%
\begin{tabular}{@{}lllccc@{}}
\toprule
Backbone & Variant & Direction & RAW-FC & LIAR-RAW & AVeriTeC \\ 
\midrule
\multirow{8}{*}{LLaMA-2} 
           & \multirow{4}{*}{self} & -> style|fact  & 61.81 & 43.06 & 84.61 \\ 
           &       & -> fact       & \good{58.06} & \good{42.91} & \good{83.35} \\ 
           &       & -> base        & \good{60.67} & \good{42.70} & \good{83.35} \\ 
           &       & -> sft         & \good{60.67} & \bad{43.33} & \good{83.35} \\ 
           \cmidrule(l){2-6} 
           & \multirow{4}{*}{cross} & -> style|fact  & 64.99 & 42.77 & - \\ 
           &       & -> fact        & \good{61.66} & \bad{43.95} & - \\ 
           &       & -> base        & \good{64.47} & \good{42.93} & - \\ 
           &       & -> sft         & \good{64.47} & \good{42.85} & - \\ 
\midrule
\multirow{8}{*}{Qwen-3} 
           & \multirow{4}{*}{self} & -> style|fact & 58.86 &46.53 & 88.21 \\ 
           &       & -> fact       & \good{58.79} & \bad{46.73} & \good{88.02} \\ 
           &       & -> base        & \bad{58.88}  & \bad{46.64} & \good{88.02} \\ 
           &       & -> sft         & \good{57.86}  & \bad{46.79} & \good{87.51} \\ 
           \cmidrule(l){2-6} 
           & \multirow{4}{*}{cross} & -> style|fact & 59.39 & 47.13 & - \\ 
           &       & -> fact       & \good{57.85} & \good{46.86} & - \\ 
           &       & -> base        & \good{58.35} & \good{46.57} & - \\ 
           &       & -> sft         & \good{57.85} & \good{46.57} & - \\ 
\bottomrule
\end{tabular}
}
\label{tab: md}
\end{table}


For model directions, we take \textsc{REFLEX} after S-EGS as the baseline (denoted as style$|$fact), and test three additional settings on Dual w/exp variants: anchoring (1) fully on fact (positives: correct verdicts, negatives: incorrect ones), (2) on the base model (positives: backbone outputs, negatives: SFT outputs), and (3) on the SFT model (the reverse of (2)). As shown in Table~\ref{tab: md}, the blue regions are dominated by red ones, further motivating our disentanglement design. Most blue regions appear on LIAR-RAW, likely due to the recency bias discussed in Section~\ref{sec:obb}.

\begin{figure}[htbp!]
\centering
\includegraphics[width=\linewidth]{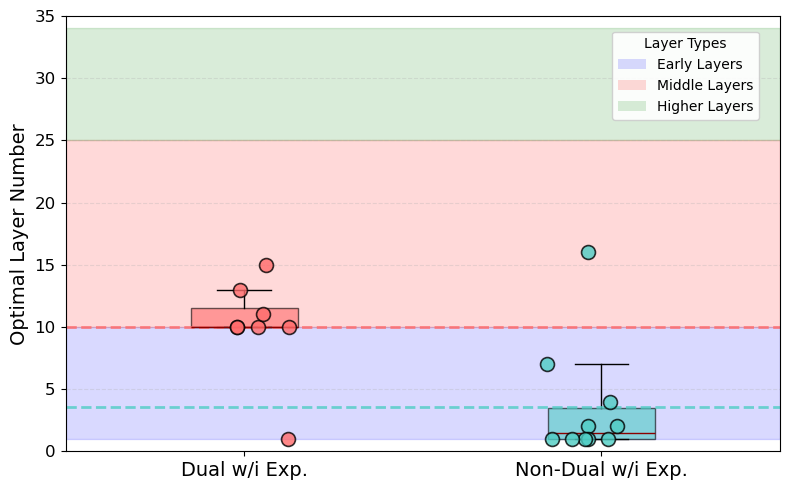} 
\caption{Optimal layers for improving different pair constructions across transformer layers. Square brackets denote optional components.}
\label{fig: layer}
\end{figure}

For optimal layers, we present the distribution of disentangled vectors across the two backbone models. Figure~\ref{fig: layer} shows that, for Dual w/o exp pairs, the largest divergences emerge in early layers (1--5), whereas for pairs with full explanations, divergences peak in the \textbf{middle layers} (10--20). This aligns with prior findings on transformer interpretability: early layers capture lexical and topical signals, middle layers encode stylistic and syntactic patterns, and higher layers capture more conceptual information~\cite{yun2021transformer}. Interestingly, this middle-layer dominance appears across both subjective styles (e.g., sycophancy and myopic reward) and objective fact-related phenomena (e.g., factuality vs.\ hallucination), and from common misconceptions to fine-grained fact-checking misinformation. This pattern suggests that contrastive signals mainly capture surface-level stylistic patterns of reasoning or knowledge representation.

For the case studies, we render cosine similarities between unrefined output tokens and optimal vectors in HTML. Red tokens denote alignment with the optimal vector direction, while blue tokens denote the opposite. As shown in Figure~\ref{fig:high-density}, red regions tend to capture correct verdict-related content, whereas blue regions are dominated by noisy or redundant syntax patterns. This further supports that \textsc{REFLEX} disentangles surface-level veracity signals from explanation style.

\section{Automatic Evaluations Results for Ablative Studies}
\label{app: ae-on-exp}
The automatic evaluation results for explanations of improved variants are shown in Table~\ref{tab: eq-all}. We also computed the standard errors of the decreases/gains for each dimension---Misleadingness, Informativeness, Soundness, and Readability---which are 0.0325, 0.0160, 0.0127, and 0.0189, respectively. Since the standard error for the decrease in Misleadingness is relatively high, and this dimension is particularly important, we further introduce pairwise human evaluation to compensate for this instability.

\begin{table*}[htb]
\caption{Explanation quality for all improved variants. Red background denotes improvement, while blue denotes decline. The omitted bar indicates that Macro-F1 did not improve in Tables~\ref{tab: bb} and~\ref{tab: pc}.}
\vspace{-10pt}
\centering
\begin{adjustbox}{max width=\textwidth}
\begin{tabular}{@{}llllllllllllll@{}}
\toprule
\multicolumn{1}{c}{} & \multicolumn{1}{c}{} & \multicolumn{4}{c}{RAW-FC} & \multicolumn{4}{c}{LIAR-RAW} & \multicolumn{4}{c}{AVeriTeC} \\
\cmidrule(lr){3-6} \cmidrule(lr){7-10} \cmidrule(lr){11-14}
\multicolumn{1}{c}{\multirow{-2}{*}{Backbone}} &
\multicolumn{1}{c}{\multirow{-2}{*}{Pair}} &
M$\downarrow $ & I & S & R &
M$\downarrow $ & I & S & R &
M$\downarrow $ & I & S & R \\
\midrule

\multirow{5}{*}{LLaMA-2-7b} & SFT
& 1.90 & 4.78 & 4.82 & 4.55
& 1.90 & 4.48 & 4.60 & 4.65
& 1.21 & 4.61 & 4.89 & 4.86 \\

& Dual - cross
& \bad{2.00} & \good{4.89} & \good{4.83} & \good{4.81}
& -- & -- & -- & --
& -- & -- & -- & -- \\

& Dual - self
& \bad{1.91} & \good{4.80} & \bad{4.77} & \good{4.75}
& \good{1.80} & \good{4.50} & \good{4.63} & \good{4.83}
& \good{1.18} & \good{4.63} & 4.89 & \good{4.88} \\

& Single - vertical
& \bad{1.95} & \good{4.87} & \good{4.84} & \textbf{\good{4.86}}
& \good{1.77} & \good{4.58} & \good{4.66} & \good{4.83}
& \good{1.18} & \good{4.65} & \bad{4.86} & \good{4.89} \\

& Single - horizontal
& \good{1.79} & \good{4.88} & \good{4.83} & \good{4.80}
& \good{1.77} & \good{4.54} & \good{4.67} & \good{4.84}
& -- & -- & -- & -- \\

\midrule

\multirow{5}{*}{Qwen-3-8b} & SFT
& 1.89 & 4.74 & 4.80 & 4.32
& 1.99 & 4.43 & 4.55 & 4.22
& 1.10 & 4.67 & 4.89 & 4.89 \\

& Dual - cross
& \good{1.83} & \good{4.87} & \bad{4.75} & \good{4.82}
& \good{1.83} & \good{4.53} & \good{4.64} & \textbf{\good{4.83}}
& -- & -- & -- & -- \\

& Dual - self
& \good{1.87} & \good{4.89} & \good{4.81} & \good{4.81}
& -- & -- & -- & --
& \bad{1.11} & \bad{4.63} & 4.89 & \good{4.90} \\

& Single - vertical
& 1.89 & \good{4.89} & \good{4.83} & \good{4.75}
& \good{1.80} & \good{4.55} & \good{4.63} & \good{4.82}
& 1.10 & 4.67 & \good{4.91} & \good{4.90} \\

& Single - horizontal
& -- & -- & -- & --
& \good{1.84} & \good{4.54} & \good{4.63} & \good{4.82}
& \bad{1.13} & \good{4.70} & \good{4.92} & \good{4.92} \\

\bottomrule
\end{tabular}
\end{adjustbox}
\label{tab: eq-all}
\end{table*}

\section{More Analysis of Statistical Correlations}
\label{app: ana-cm}

As shown in Figure~\ref{fig:cm}, three findings emerge:
\textbf{(1)} Performance is primarily driven by \textbf{faithfulness-related metrics}: both F-score and accuracy are strongly negatively correlated with misleadingness (-0.85, -0.96) and strongly positively correlated with soundness (0.91, 0.78).
\textbf{(2)} Readability plays a secondary but \textbf{consistent role}: it correlates positively with both F-score (0.67) and accuracy (0.81), suggesting that clearer explanations tend to align with better predictions.
\textbf{(3)} Informativeness exhibits \textbf{a clear trade-off} with effectiveness. It is negatively correlated with readability (-0.35) and accuracy (-0.04), while positively correlated with both misleadingness (0.26) and soundness (0.57), indicating that adding more background information can make explanations appear more logically grounded, but often introduces additional noise that reduces clarity and hurts decision quality.

\section{Case Studies}
\label{app: case-study}
\textsc{REFLEX} not only corrects cases in which both the explanation and the verdict are completely wrong, as shown below, but also reduces inconsistencies with the verdict and self-contradictions within explanations. These issues are often coupled, but can also appear independently. Manual case studies further show that \textsc{REFLEX} reduces irrelevant information and overconfident hallucinations in explanations.

\begin{myquote}

[Claim] 

Canada would like to reopen the border with the U.S.

[Evidences]

(1) When were border crossing restrictions between Canada and the USA first introduced? The Prime Minister, Justin Trudeau, announced on March 18 2020 that Canada and the United States have agreed to temporarily restrict all non-essential travel across the Canada-U.S. border. The new measures will take effect at 12:01 a.m. EDT on Saturday, March 21, 2020, and end at 12:00 p.m. EDT on Tuesday, April 21, 2020. At that time, Canada and the United States will review the measures. The Prime Minister, Justin Trudeau, announced on March 18 2020 that Canada and the United States have agreed to temporarily restrict all non-essential travel across the Canada-U.S. border. The new measures will take effect at 12:01 a.m. EDT on Saturday, March 21, 2020, and was set to end at 12:00 p.m. EDT on Tuesday, April 21, 2020.  

(2) Were Canadian/American border crossings still restricted on 18/09/2020? Yes.  

(3) Why were Canada/USA border restrictions extended on 18/09/2021? The Government of Canada 2019s top priority remains the health and safety of Canadians, and this extension will continue to protect people on both sides of the border while ensuring the flow of essential goods and services between our two countries. In order to limit the further spread of coronavirus. Canada's Health Minister Patty Hajdu mentioned in an interview that she was \"happy\" with the border closure because Canada already has its \"hands full\" with the pandemic. The Government of Canada 2019s top priority remains the health and safety of Canadians, so this extension will continue to protect people on both sides of the border.  

(4) Were Canadian public happy that the boarder was still closed? 85\% of Canadians would like the borders with the U.S. to remain closed until at least the end of 2020. 

[Label]

REFUTED
\newline \newline
[SFT]

from AVeriTeC, Qwen-3, Dual-wi/exp

[Anonymous as B in Judgement-1 and swapped as A in Judgement-2]

Verdict: SUPPORTED. Explanation: The assistant incorrectly labels the claim as supported, despite the evidence showing that the border was closed and most Canadians wanted it to stay closed. This misrepresents the claim's veracity.
\newline \newline
[REFLEX] KV-layer10-multiplier1.5

[Anonymous as A in Judgement-1 and swapped as B in Judgement-2]

Verdict: REFUTED. Explanation: The evidence shows that the Canada-U.S. border remained closed and that the majority of Canadians preferred it to stay closed. Therefore, the claim that Canada would like to reopen the border is contradicted and refuted.
\newline \newline
Judgment-1: Assistant A correctly identifies that the claim is refuted, aligning with the true label. Assistant B incorrectly labels it as supported, which is misleading because it \textbf{contradicts the evidence}. Therefore, Assistant B's explanation is more misleading.

Judgment-2: Both assistants discuss the border status and public opinion. The claim is that Canada would like to reopen the border, which is directly contradicted by the evidence. Assistant A correctly labels the claim as refuted, while Assistant B incorrectly supports it. Assistant B's explanation \textbf{misrepresents the evidence} and is therefore more misleading.

\end{myquote}

\subsection{Misleading Verdict Inconsistency}

\begin{myquote}

[Claim]

Pre-existing conditions are covered under my (health care) plan.

[Label]

FALSE
\newline \newline
[SFT]

from LLaMA-2, Liar-RAW, Dual wi/exp.-self 

[Anonymous as B in Judgement-1 and swapped as A in Judegement-2]

Verdict: FALSE. Explanation: But the plan \textcolor{red}{doesn't cover} pre-existing conditions. It \textcolor{red}{only covers} pre-existing conditions that existed before the plan was purchased.
\newline \newline
[REFLEX] KV-layer1-multiplier1.5

[Anonymous as A in Judgement-1 and swapped as B in Judegement-2]

Verdict: FALSE. Explanation: But the law does not require them to do so. And the law specifically excludes coverage of pre-existing conditions before the effective date of the new plan.
\newline \newline
Judgment-1: Both assistants correctly identify the claim as FALSE, which aligns with the true label. However, Assistant A's explanation clearly states that the law does not require coverage of pre-existing conditions and that such conditions are specifically excluded before the effective date of the new plan. This is a straightforward and accurate explanation supporting the false verdict. Assistant B's explanation is \textcolor{red}{confusing and somewhat contradictory}. It says the plan \"doesn't cover pre-existing conditions,\" which is correct, but then adds \"It only covers pre-existing conditions that existed before the plan was purchased,\" which contradicts the first part and is misleading because it implies some coverage of pre-existing conditions, which is not true. Therefore, Assistant B's explanation is more misleading due to the contradictory and unclear statement about coverage.

Judgment-2: Both assistants correctly identify the claim as FALSE, which aligns with the true label. However, Assistant A's explanation is somewhat \textcolor{red}{confusing and contradictory}: it states the plan \"doesn't cover pre-existing conditions,\" but then says \"It only covers pre-existing conditions that existed before the plan was purchased,\" which is logically inconsistent and misleading. Assistant B's explanation is clearer and more accurate, noting that the law does not require coverage of pre-existing conditions and that such conditions are excluded before the plan's effective date, which aligns with typical health insurance policies. Therefore, Assistant A's explanation is more misleading due to its contradictory phrasing.
\end{myquote}

\subsection{Intra-Explanation Self-Contradictions}

\begin{myquote}

[Claim]

Health care reform \"would \textbf{bring down the deficit} by as much as \$1 trillion over the next two decades.

[Label]

HALF-TRUE
\newline \newline
[SFT]

from Liar-Raw, Qwen-3, Dual-wi/exp

[Anonymous as B in Judgement-1 and swapped as A in Judegement-2]

Verdict: HALF-TRUE. Explanation: Obama said health care reform "would bring down the deficit by as much as \$1 trillion over the next two decades." The CBO has estimated that the health care bill would \textcolor{red}{reduce the deficit} by \$1 trillion over 10 years. But the CBO has also said that the bill would \textcolor{red}{increase the deficit by \$100 billion over the next 10 years}. Obama’s statement is partially accurate but leaves out important details.
\newline \newline
[REFLEX] KV-layer15-multiplier-1.5

[Anonymous as A in Judgement-1 and swapped as B in Judegement-2]

Verdict: HALF-TRUE. Explanation: Obama said health care reform "would bring down the deficit by as much as \$1 trillion over the next two decades." The CBO has said that the health care bill would reduce the deficit by \$143 billion over 10 years. The CBO has also said that the health care bill would reduce the deficit by \$1.7 trillion over 20 years. Obama’s statement is partially accurate but leaves out important details.
\newline \newline
Judgment-1: Assistant A provides CBO estimates (\$143 billion over 10 years, \$1.7 trillion over 20 years) that support the claim being HALF-TRUE, noting omitted details. Assistant B presents \textcolor{red}{contradictory information} (\$1 trillion reduction over 10 years and a \$100 billion increase over 10 years), which is confusing and misleading. Therefore, Assistant B's explanation is more misleading.

Judgment-2: Both assistants label the claim as HALF-TRUE. However, Assistant A contains a \textcolor{red}{contradictory statement} about CBO estimates (both \$1 trillion reduction and \$100 billion increase over 10 years), which could mislead the reader. Assistant B provides consistent figures (\$143 billion over 10 years, \$1.7 trillion over 20 years) that align better with the claim. Therefore, Assistant A's explanation is more misleading.
\end{myquote}

\subsection{Intra-Explanation Irrelevant Information}

\begin{myquote}

[Claim]

U.S. actor Tom Hanks is related to "Mister Rogers' Neighborhood" host Fred Rogers.

[Label]

TRUE
\newline \newline
[SFT]

from Qwen-3, RAW-FC, Dual-wi/exp-self

[Anonymous as B in Judgement-1 and swapped as A in Judegement-2]

Verdict: TRUE. Explanation: In June 2020, as the world mourned the death of Fred Rogers, host of the long-running children’s television show "Mister Rogers’ Neighborhood," a rumor started to circulate on social media that the late entertainer was related to actor Tom Hanks.
The rumor was largely spread by a Facebook user who posted a meme containing a photograph of Hanks and Rogers along with the caption: “Did you know that Tom Hanks is related to Fred Rogers?” The claim that Hanks and Rogers were related was also spread by a number of other social media users, as well as in a viral tweet by the comedian Patton Oswalt. The claim that Hanks and Rogers were related is true. \textcolor{red}{Hanks’ paternal grandmother, Lucille, was the sister of Rogers’ paternal grandmother, Marion.}. Hanks confirmed the familial connection in a 2018 interview with the "Today" show. Hanks said that he first learned about the connection when he was a child. He also said that he was a big fan of Rogers’ show and that he would watch it with his sister. “I was a big fan of Mr. Rogers,” Hanks said. “I loved that show. I would watch it with my sister. I knew that we were related. I knew that we were related. I knew that we were related.” \textcolor{red}{Hanks’ paternal grandmother, Lucille, was the sister of Marion, the mother of Fred Rogers. Hanks’ great-great-grandmother, Mary, was the sister of Rogers’ great-great-grandmother, Ann.} The two men were related by blood, but they were also connected by their careers. Hanks played the role of “Mr. Hand” in the 2003 film “A Walk to Remember.” In the movie, “Mr. Hand” is a children’s television host who bears a passing resemblance to Rogers. Hanks also lent his voice to the character of “Mr. Hand” in the 2015 animated film “The SpongeBob Movie: Sponge Out of Water.” \textcolor{red}{Here’s a clip from “A Walk to Remember” featuring Hanks as “Mr. Hand”: Fred Rogers passed away on Feb. 27, 2003. Tom Hanks is still alive and well.} 
\newline \newline
[REFLEX] IV-layer10-multiplier1.5

[Anonymous as A in Judgement-1 and swapped as B in Judegement-2]

Verdict: TRUE. Explanation: In June 2020, as the world mourned the death of Fred Rogers, host of the long-running children’s television show "Mister Rogers’ Neighborhood," a rumor started to circulate on social media that the late entertainer was related to actor Tom Hanks.
The claim was made in a tweet posted by the "Mister Rogers’ Neighborhood" Facebook page, which stated that Rogers was Hanks’ “great-great-great uncle.” The claim was also shared in a meme posted to the "Mister Rogers’ Neighborhood" Instagram page. The claim that Hanks and Rogers were related was true. Hanks’ mother, Nancy, was Rogers’ first cousin, once removed. Hanks confirmed the familial connection in a 2018 interview with the "Mister Rogers’ Neighborhood" Facebook page. Hanks said that he first learned about the connection when he was 10 years old. “I was 10 years old and I was watching ‘Mister Rogers’ Neighborhood,’” Hanks said. “And I was like, ‘Who is this guy?’ And then my mom said, ‘That’s your great-great-uncle.’ And I was like, ‘What?’ And then she said, ‘Yeah, he’s your great-great-uncle.’ And I was like, ‘Wow.’” Hanks’ mother, Nancy, was the daughter of Rogers’ first cousin, Mary, and Rogers’ uncle, James. Hanks’ mother passed away in 2010. In a 2018 interview with the "Mister Rogers’ Neighborhood" Facebook page, Hanks said that he and Rogers never met, but that he always felt a connection to the entertainer. “I always felt like I knew him,” Hanks said. “I always felt like I knew him. And I think that’s because I knew my mom. And I knew my mom loved him. And I knew my mom loved what he stood for. And I knew my mom loved what he did. And I knew my mom loved the way he made people feel.”
\newline \newline
Judgment-1: Both assistants correctly identify the claim as TRUE, which aligns with the true label. However, Assistant A provides a clear and accurate explanation of the familial relationship, stating that Tom Hanks' mother was Fred Rogers' first cousin once removed, and includes a direct quote from Hanks confirming the connection. This explanation is consistent with known facts and does not introduce unrelated or incorrect information.
Assistant B, while also concluding the claim is true, presents a \textcolor{red}{confusing and inaccurate} family relationship, stating that Hanks' paternal grandmother was the sister of Rogers' paternal grandmother, and further complicates the genealogy with additional, unsupported claims. Moreover, Assistant B \textcolor{red}{introduces irrelevant information} about Hanks playing a character named "Mr. Hand" in unrelated films, which is misleading and unrelated to the claim about their familial connection. This distracts from the factual basis and could mislead the reader about the nature of their relationship.
Therefore, Assistant B's explanation is more misleading due to inaccuracies in the family relationship and the inclusion of irrelevant and incorrect details.

Judgment-2: Both Assistant A and Assistant B correctly identify the claim as TRUE, which aligns with the true label. However, Assistant A provides a detailed but somewhat confused and \textbf{inconsistent explanation of the familial relationship}, mentioning multiple relatives and roles that do not clearly establish the connection. The explanation also includes \textbf{irrelevant information} about Tom Hanks' acting roles, which could mislead the reader about the nature of the relationship. Assistant B, on the other hand, offers a clearer and more straightforward explanation of the family connection, specifying the relationship through Hanks' mother and providing a plausible narrative of how Hanks learned about it. Although some details in Assistant B's explanation might not be fully verifiable, it stays focused on the familial link without introducing unrelated information.
Therefore, Assistant A's explanation is more misleading due to its confusing and extraneous details that do not clearly support the claim.
\end{myquote}

\subsection{Intra-Explanation Overconfident Hallucinations}

\begin{myquote}

[Claim]

24 of every 100 kids in US schools speak Spanish.

[Evidences]

(1) What percentage of kids in US schools speak Spanish? The percentage of public school students in the United States who were English learners was 10.4\% in the fall of 2019 and 9.2\% in the fall of 2010. About 6 in 10 young Hispanics say they use Spanish.

[Label]

Not Enough Evidence
\newline \newline
[SFT]

from LLaMA-2, AVeriTeC, Dual wi/exp 

[Anonymous as B in Judgement-1 and swapped as A in Judegement-2]

Verdict: Refuted. Explanation: The answer and source show that the amount of children that speak Spanish is \textcolor{red}{less than 24 in 100} so the claim is refuted. 
\newline \newline
[REFLEX] KV-layer11-multiplier-1.5

[Anonymous as A in Judgement-1 and swapped as B in Judegement-2]

Verdict: The answer and source show that the statistic is wrong so the claim is refuted.
\newline \newline
Judgment-1: Both Assistant A and Assistant B conclude that the claim is refuted, stating that the evidence shows the percentage of kids speaking Spanish is less than 24\%. However, the true label is "Not Enough Evidence," indicating that the provided evidence does not sufficiently confirm or deny the claim. Both assistants incorrectly assert a refutation, but Assistant B's explanation is slightly more misleading because it \textcolor{red}{confidently quantifies} the percentage as "less than 24 in 100," implying a definitive comparison that the evidence does not fully support. Assistant A's explanation is more general, simply stating the statistic is wrong without specifying the extent. Therefore, Assistant B's explanation deviates more from the true label.

Judgment-2: Both Assistant A and Assistant B conclude that the claim is refuted, which is inconsistent with the true label "Not Enough Evidence." The evidence provided does not directly state the percentage of all kids in US schools who speak Spanish; it only provides data on English learners and the proportion of young Hispanics who use Spanish, which is insufficient to confirm or refute the claim definitively. Assistant A attempts to quantify the refutation by stating the number is less than 24 in 100, which is an \textcolor{red}{assumption not fully supported by the evidence}. Assistant B simply states the statistic is wrong without elaboration. Both explanations mislead by asserting a refutation when the evidence is insufficient to do so. However, Assistant A's explanation is more misleading because it implies a specific numeric conclusion that is not supported, whereas Assistant B's explanation is more general.
\end{myquote}





\end{document}